\PassOptionsToPackage{table}{xcolor}
\documentclass[manuscript,screen]{acmart}

\AtBeginDocument{%
  }

\setcopyright{none}
\copyrightyear{2026}
\acmYear{2026}
\acmDOI{}
\usepackage{makecell}
\usepackage{multirow}
\usepackage{multicol}
\usepackage[normalem]{ulem}
\useunder{\uline}{\ul}{}
\usepackage{hyperref}
\usepackage{threeparttable}
\usepackage{algorithm}
\usepackage{algpseudocode}
\usepackage{amsmath,amsfonts}
\usepackage{amsthm}
\newtheorem{definition}{Definition}
\usepackage{float}
\usepackage{enumitem}
\usepackage{graphicx}
\usepackage{subcaption}
\usepackage{afterpage}

\usepackage{caption}
\usepackage{dblfloatfix}
\usepackage{booktabs} 
\usepackage{afterpage}
\usepackage{cleveref}
\usepackage{svg}
\usepackage{wrapfig}

\colorlet{blue}{black}
\newcommand{\revisiontext}[1]{{\color{blue}#1}}
\begin{document}
\setlength{\emergencystretch}{1em}

\title{From XXLTraffic to EvoXXLTraffic: Scaling Traffic Forecasting to Sensor-Evolving Networks}


\author{Du Yin}
\email{du.yin@unsw.edu.au}
\affiliation{%
  \institution{University of New South Wales} 
  \city{Sydney}
  \state{NSW}
  \country{Australia}  
}

\author{Hao Xue}
\email{hao.xue1@unsw.edu.au}
\affiliation{%
  \institution{University of New South Wales} 
  \city{Sydney}
  \state{NSW}
  \country{Australia}  
}

\author{Arian Prabowo}
\email{arian.prabowo@unsw.edu.au}
\affiliation{%
  \institution{University of New South Wales} 
  \city{Sydney}
  \state{NSW}
  \country{Australia}  
}

\author{Shuang Ao}
\email{shuang.ao@unsw.edu.au}
\affiliation{%
  \institution{University of New South Wales} 
  \city{Sydney}
  \state{NSW}
  \country{Australia}  
}

\author{Flora Salim}
\email{flora.salim@unsw.edu.au}
\affiliation{%
  \institution{University of New South Wales} 
  \city{Sydney}
  \state{NSW}
  \country{Australia}  
}

\renewcommand{\shortauthors}{Du Yin et al.}

\begin{abstract}
\revisiontext{Traffic forecasting benchmarks commonly fix the sensor set, although deployed sensing networks change their coverage over time. We present EvoXXLTraffic, a sensor-evolving extension of the XXLTraffic dataset family, with yearly sensor catalogs, persistent station IDs, flow matrices, and geographic graph snapshots across nine California PeMS districts, spanning up to 27 years. The benchmark separates all-sensor forecasting from supervised adaptation to sensors absent from the preceding year. We evaluate 13 baseline adaptations on PEMS04 with five seeds and PEMS05 with three seeds, and seven of them on the other seven districts with three seeds. Each yearly task uses 31 stored days, explicit observation masks, and chronological evaluation. Under the frozen configurations, PECPM has the lowest full-horizon mean MAE on both all-sensor and new-node populations on PEMS04, while \mbox{Online-AN} leads both populations on PEMS05. Among the seven methods run on every district, \mbox{Online-AN} has the lowest all-sensor mean MAE in eight of the nine districts, and \mbox{Online-NN} has the lowest new-node mean MAE in six. A matched control on PEMS05 and PEMS06 finds that training on the evolving catalog changes mean MAE on the sensors present in every 2018--2025 catalog by less than 1\% in either direction. Method ordering depends on district, evaluation population, horizon, and metric, and seed variation does not establish statistical significance. The benchmark documents implementation adaptations and uses retrospective geographic graphs, limiting claims about original-paper rankings and prospective deployment. Inherited fixed-sensor gap and hourly/daily tasks provide complementary temporal settings. Code and results: \url{https://github.com/cruiseresearchgroup/TSAS26-EvoXXLTraffic}. Data: \url{https://huggingface.co/datasets/yind7/EvoXXLTraffic}.}
\end{abstract}



\maketitle
\section{Introduction}
\label{sec:intro}

\revisiontext{Traffic forecasting supports transport monitoring and planning by predicting measurements at instrumented locations. Existing benchmarks often define one fixed sensor set and evaluate adjacent observation and prediction windows. Two different changes complicate longer-term use: traffic distributions can shift at existing locations, and the set of instrumented locations can change. A fixed sensor intersection supports comparison over time but excludes locations that enter or leave the sensor set during the period. Geographic proximity among sensors also need not coincide with directed road connectivity.}

\begin{figure*}[t]
\centering
\includegraphics[width=0.99\linewidth]{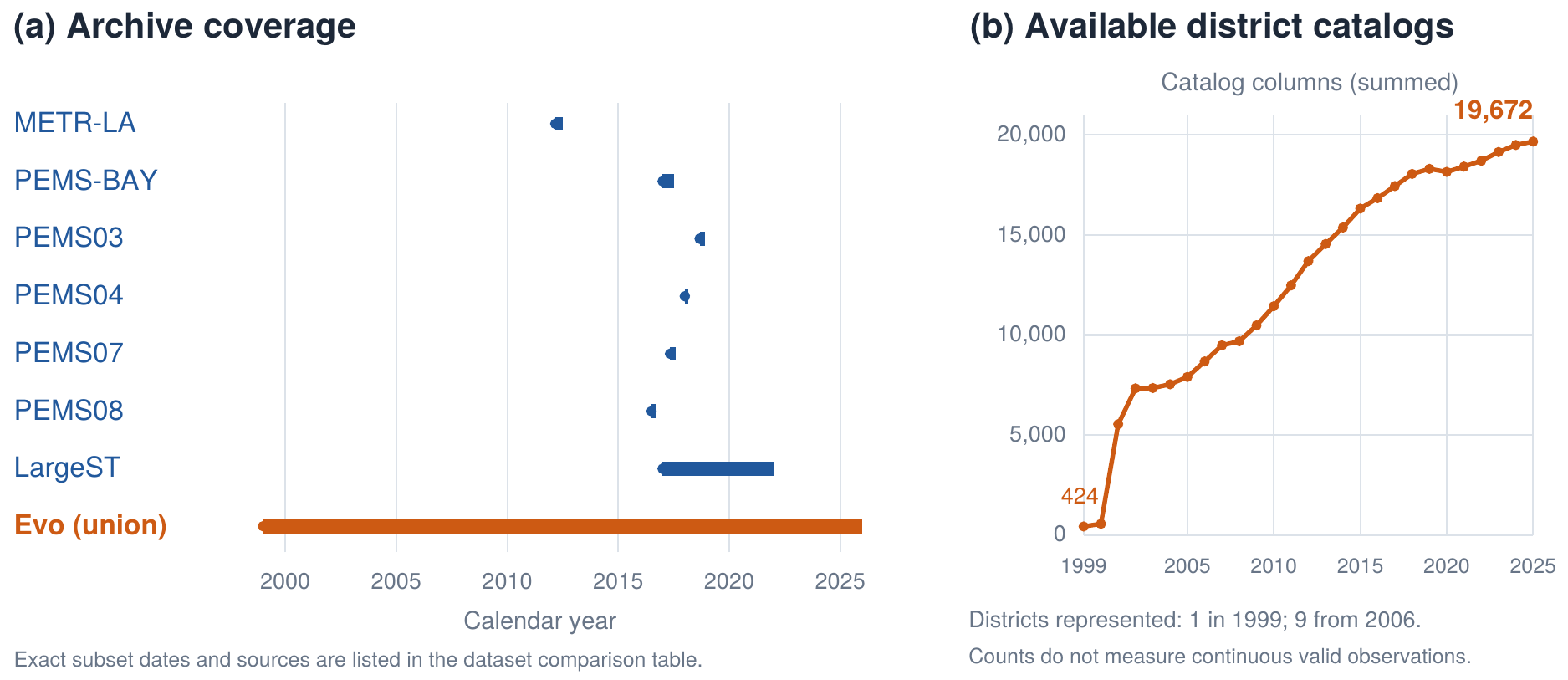}
\caption{\revisiontext{Archive coverage and changing catalog size. Left: representative fixed-sensor datasets and the union of EvoXXLTraffic district coverage, with sources and exact subset dates in Table~\ref{tab:data1}. Right: sums of the stored yearly catalogs across available districts. The number of represented districts changes from one to nine, so these totals are not a fixed-region growth rate. PEMS04 retains 2001 in the archive but enters the benchmark in 2002. Counts describe catalog membership, not physical road connectivity or complete observations.}}
\Description{Overview of temporal archive coverage and changing yearly sensor populations.}
\label{fig:intro}
\end{figure*}

\revisiontext{Our SIGSPATIAL 2025 work introduced XXLTraffic, a dataset family for fixed-sensor forecasting across multi-year gaps and at hourly/daily resolutions~\cite{yin2025xxltraffic}. This article extends that resource with EvoXXLTraffic: yearly PeMS sensor catalogs, persistent station IDs, flow matrices, and graph snapshots. Figure~\ref{fig:intro} motivates the extension through temporal coverage and sensor-set evolution. When the sensor set evolves, a method must specify which sensor identities it can forecast, how it reuses historical state, and how errors on newly entering sensors relate to errors on the current catalog. Sensor-count growth reflects recorded sensing coverage and is not a measurement of road-network growth.}

\subsection{Research Questions and Benchmark Scope}
\label{subsec:research_questions}
\revisiontext{We organize the sensor-evolving benchmark around three questions. RQ1: under a common chronological evaluation protocol, how do retraining, online adaptation, and continual forecasting compare on the current sensor catalog? RQ2: does that comparison change when evaluation is restricted to sensors absent from the preceding year? RQ3: how stable are these observations across random seeds, districts, and a recent calendar-period slice? These questions concern benchmark behavior, and we do not introduce a new forecasting model.}

\revisiontext{Yearly tasks provide a reproducible unit for aligning sensor identities, graph snapshots, and checkpoint transitions over a long archive. Within each year, the benchmark uses the first 31 stored days with chronological training, validation, and test splits. This is a bounded experimental abstraction, not continuous full-year deployment or event-by-event sensor arrival. Current-year training includes new sensors, so the task is supervised new-node adaptation rather than zero-shot cold start. Reporting both current-catalog and new-node errors distinguishes maintaining coverage from improving error on a fixed population. Whether training on the evolving catalog changes accuracy on long-lived sensors is tested separately against a matched fixed-catalog control (Section~\ref{subsec:fixed_vs_evolving}).}

\revisiontext{Dataset coverage and experimental coverage are distinct: EvoXXLTraffic contains nine districts, and the benchmark compares all 13 methods on PEMS04 over 2002--2025 with seeds 47--51 and on PEMS05 over 2005--2025 with seeds 47--49. Seven of these methods are also run on the other seven districts with seeds 47--49, so every district has a corrected comparison. Under the frozen configurations, PECPM leads full-12-lead mean MAE for both all-sensor and new-node populations on PEMS04, while Online-AN leads both on PEMS05. Among the seven methods, Online-AN has the lowest all-sensor mean in eight of the nine districts. Online-NN trails it on all sensors in every district but has the lowest new-node mean in six, so we report each evaluation population and its variation rather than a universal winner.}

\subsection{Extension over the Conference Version}
\label{subsec:journal_extension}
\revisiontext{The fixed-sensor gap, hourly, and daily settings are inherited from the conference work~\cite{yin2025xxltraffic} and are not counted as new contributions. The journal extension adds the following components:}
\begin{itemize}
\item \revisiontext{\textbf{Data organization:} yearly sensor catalogs with persistent IDs, yearly flow matrices, and graph snapshots across nine PeMS districts, with the archive extended through 2025.}
\item \revisiontext{\textbf{Evaluation protocol:} sequential yearly tasks with checkpoint provenance, explicit observation masks, chronological testing, and separate all-sensor, new-to-previous-year, first-ever, retained, and returning cohorts.}
\item \revisiontext{\textbf{Benchmark evidence:} 13-method comparisons on PEMS04 and PEMS05 plus seven-method comparisons on the other seven districts, from 251 completed method--seed runs. Three paired coordinate-policy sensitivity runs and 24 runs of a matched fixed- versus evolving-catalog training control bring the total to 278 runs. These are adaptations of existing baselines, not newly proposed forecasting algorithms.}
\item \revisiontext{\textbf{Analysis:} corrected catalog statistics and coordinate-fallback counts, seed variability, differences between evaluation cohorts, a recent-period diagnostic, fixed-catalog coverage with a matched fixed-catalog training control, three-seed coordinate-policy sensitivity, and end-to-end runtime reporting. We separate retrospective graph assumptions from prospective deployment claims.}
\end{itemize}
\revisiontext{Our comparison with the conference version uses its public author version, which carries the SIGSPATIAL 2025 DOI (arXiv:2406.12693v3). That paper already motivates growing sensing networks, describes the long archive, and presents the gap, hourly, and daily experiments. We therefore do not claim sensor growth itself, the original archive, or those experiments as new. The additions here operationalize changing sensor populations through yearly identity alignment, a masked sequential protocol, and the nine-district evidence described above. Sections~\ref{subsec:evoxxltraffic}, \ref{subsec:evo_streaming_protocol}, and~\ref{exp_section6} and Appendix~\ref{app:evo_details} are new. Sections~\ref{subsec:xxltraffic_fixed}, \ref{subsec:data_overview}, and~\ref{subsec:fixed_protocols} and Appendix~\ref{app:legacy_results} summarize the inherited fixed-sensor material.}

\revisiontext{The scientific purpose of the extension is to make changing forecast populations observable and testable. Fixed-dimensional benchmarks cannot directly reveal whether an apparently strong average hides different behavior on retained and newly entering sensors. EvoXXLTraffic makes this comparison possible while preserving sensor identities and documenting the limits of the graph and observation construction.}

\section{Related Work and Dataset Comparison}

\subsection{Traffic Forecasting and Dataset Coverage}
\revisiontext{Traffic forecasting methods combine temporal modeling with spatial information in different ways. Examples include graph convolution in STGCN~\cite{yu2018spatio}, recurrent diffusion modeling in DCRNN~\cite{li2018diffusion}, and attention in ASTGCN~\cite{guo2019attention}. Multivariate forecasting methods such as LSTNet~\cite{lai2018modeling} and Informer~\cite{zhou2021informer} also evaluate longer prediction windows. Forecast horizon, temporal sampling interval, archive duration, and sensor-set evolution are separate properties: a long archive does not imply that a model predicts several years ahead, and a longer output window does not itself introduce new sensors.}

\revisiontext{Common fixed-sensor datasets cover a few months, whereas LargeST provides five years and up to 8,600 sensors~\cite{liu2024largest}. Table~\ref{tab:data1} compares representative released subsets using their stated coverage, rather than conflating them with the complete PeMS districts. XXLTraffic adds a longer archive and fixed-sensor gap/hourly/daily tasks~\cite{yin2025xxltraffic}, and EvoXXLTraffic adds yearly catalogs and identity-aware evaluation across changing forecast populations. Its contribution lies in this organization and evaluation scope rather than in the size of any single district.}

\subsection{Evolving Sensor Sets and Continual Forecasting}
\revisiontext{TrafficStream~\cite{ijcai2021p0498} studies streaming traffic forecasting with graph neural networks and continual learning. PECPM~\cite{wang2023pattern} and STKEC~\cite{wang2023knowledge} develop expansion and consolidation mechanisms for changing graphs. Other approaches use prompt tuning (EAC~\cite{chen2025expand}), a scalable contextual pattern bank (STBP~\cite{liugeneral}), expanding-variate learning (STEV~\cite{ma2025beyond}), or pattern retrieval (STRAP~\cite{zhang2025strap}). ST-TTC~\cite{chen2025learning} studies prediction calibration during testing. These strategies differ in the state they retain and when they can use labels, and a shared benchmark must make those choices explicit.}

\revisiontext{New-location prediction also admits different supervision regimes. SCPT~\cite{prabowo2024traffic} considers forecasting at new roads through spatial pretraining, whereas the present yearly protocol gives each method current-year training observations at newly entering catalog sensors. The latter measures supervised adaptation, not zero-shot prediction. EvoXXLTraffic extends the temporal setting in which such adaptation can be evaluated. The provided graph is a geographic proximity graph, so a change in its nodes or edges does not by itself demonstrate a change in physical road connectivity.}

\section{Preliminaries and Problem Taxonomy}
\label{sec:preliminaries}

In this section, we define the forecasting settings considered in the XXLTraffic dataset family. The original XXLTraffic focuses on fixed-sensor long-horizon and gap-based forecasting, while EvoXXLTraffic extends the setting to sensor-evolving forecasting where the active sensor set and graph structure
change over calendar years.

\begin{definition}[Traffic datasets]
Traffic data consists of vehicle flow, speed, or occupancy observations
collected by sensors distributed across a traffic network. A traffic sequence is
represented as
\begin{equation}
    \mathbf{X}_{1:T}
    =
    [\mathbf{x}_1, \mathbf{x}_2, \ldots, \mathbf{x}_T]
    \in
    \mathbb{R}^{T \times N \times C},
    \label{eq:traffic_tensor}
\end{equation}
where $T$ is the number of time steps, $N$ is the number of sensors, and $C$ is
the number of traffic features. When spatial relations are used, the traffic
network is associated with an adjacency matrix
$\mathbf{A} \in \mathbb{R}^{N \times N}$.
\end{definition}

In the fixed-sensor subsets of XXLTraffic, the same sensor set is used across different years. This design supports controlled long-horizon and gap-based forecasting, because the observation and prediction windows are defined over the same sensors.

\begin{definition}[Standard traffic forecasting]
Given an input window of length $\alpha$, standard traffic forecasting predicts the next $\beta$ time steps:
\begin{equation}
    [\mathbf{x}_{t-\alpha+1}, \ldots, \mathbf{x}_{t}]
    \rightarrow
    [\mathbf{x}_{t+1}, \ldots, \mathbf{x}_{t+\beta}].
    \label{standard_forecasting}
\end{equation}
Short-term traffic forecasting usually sets $\alpha=\beta=12$ for 5-minute data, corresponding to one hour of historical observations and one hour of predictions. Long-term multivariate forecasting follows the same adjacent-window formulation but uses longer input or prediction horizons.
\end{definition}

\begin{definition}[Gap-based extremely long forecasting]
Extremely long forecasting with gaps introduces a non-contiguous temporal gap
$g$ between the observation window and the prediction window:
\begin{equation}
    [\mathbf{x}_{t-\alpha+1}, \ldots, \mathbf{x}_{t}]
    \rightarrow
    [\mathbf{x}_{t+g+1}, \ldots, \mathbf{x}_{t+g+\beta}].
    \label{eq:gap_forecasting}
\end{equation}
When $g=0$, this reduces to standard adjacent-window forecasting. When $g>0$,
the model is evaluated under long temporal distribution shifts, such as
one-year, one-and-a-half-year, or two-year gaps.
\end{definition}

\begin{definition}[Sensor-evolving traffic dataset]
The fixed-sensor setting is useful for gap-based forecasting, but real traffic
sensing systems evolve over time as new detectors are installed and existing
detectors become inactive. EvoXXLTraffic therefore represents each calendar year
with its own \revisiontext{stored sensor catalog}, traffic observations, and graph~snapshot:
\begin{equation}
    \mathcal{D}_{\mathrm{evo}}^{d}
    =
    \left\{
        \left(
            \mathbf{X}_{y}^{d},
            \mathbf{A}_{y}^{d},
            \mathcal{V}_{y}^{d}
        \right)
    \right\}_{y=y_0}^{y_1},
    \qquad
    \mathbf{X}_{y}^{d}
    \in
    \mathbb{R}^{T_y^d \times N_y^d \times C},
    \quad
    N_y^d = |\mathcal{V}_{y}^{d}|.
    \label{eq:evo_dataset}
\end{equation}
Here $d$ denotes a district, $y$ denotes a calendar year,
$\mathcal{V}_{y}^{d}$ is the \revisiontext{stored sensor catalog for year $y$, constructed from an anchor day's station IDs as described in Section~\ref{subsubsec:evo_flow_preprocessing}},
$\mathbf{X}_{y}^{d}$ is the yearly traffic tensor, and
$\mathbf{A}_{y}^{d} \in \mathbb{R}^{N_y^d \times N_y^d}$ is the yearly graph
snapshot.
\end{definition}

\begin{definition}[Sensor evolution]
Given two consecutive years $y-1$ and $y$, the newly added sensors are defined as
\begin{equation}
    \Delta \mathcal{V}_{y}^{d}
    =
    \mathcal{V}_{y}^{d}
    \setminus
    \mathcal{V}_{y-1}^{d}.
    \label{eq:new_sensors}
\end{equation}
Sensors in $\mathcal{V}_{y}^{d} \cap \mathcal{V}_{y-1}^{d}$ are \revisiontext{retained} sensors,
while sensors in $\mathcal{V}_{y-1}^{d} \setminus \mathcal{V}_{y}^{d}$ \revisiontext{are absent from the current catalog. Catalog absence alone does not establish detector removal or inactivity. The set $\Delta\mathcal{V}_y^d$ includes returning sensors that were present in an earlier year but absent in $y-1$. First-ever sensors instead belong to $\mathcal{V}_y^d\setminus\bigcup_{t<y}\mathcal{V}_t^d$, where the history begins at the benchmark's first year. We assign no change cohort to that first year. Because current-year training includes these sensors, our evaluation measures supervised new-node adaptation rather than zero-shot cold-start prediction.}
\end{definition}

\paragraph{Problem taxonomy.}
\revisiontext{The XXLTraffic dataset family supports both fixed-sensor and sensor-evolving forecasting settings. The original XXLTraffic emphasizes extremely long temporal coverage and gap-based prediction under a fixed sensor set, while EvoXXLTraffic
exposes yearly sensor growth, changing graph structures, and new-node adaptation.}

\begin{table}[h!]
    \centering
    \caption{Problem taxonomy of the XXLTraffic dataset family.}
    \label{tab:problem_taxonomy}
    \resizebox{\linewidth}{!}{
    \begin{tabular}{lcccc}
        \toprule
        \textbf{Setting}
        & \textbf{Sensor set}  & \textbf{Graph}  & \textbf{Temporal relation} & \textbf{Main challenge} \\
        \midrule
Short-term forecasting & Fixed  & Fixed & Adjacent  &  Local spatio-temporal dependency
    \\
 Long-term forecasting & Fixed  &  Fixed &  Adjacent  & Long prediction horizon
    \\
 Gap-based extremely long forecasting & Fixed &   Fixed  & Non-contiguous & Long temporal distribution shift
    \\
Sensor-evolving forecasting & Yearly evolving & Yearly evolving & Streaming & New sensors and dynamic graphs
    \\
    \bottomrule
    \end{tabular}
    }
\end{table}


\section{The XXLTraffic Dataset Family}

\subsection{XXLTraffic: Fixed-Sensor Long-Horizon Subsets}
\label{subsec:xxltraffic_fixed}

\subsubsection{Data Collection}
\label{sec:data_collection}

\revisiontext{The inherited XXLTraffic collection uses the California Department of Transportation (Caltrans) Performance Measurement System\footnote{\url{https://pems.dot.ca.gov/}} (PeMS)~\cite{chen2001freeway} and Transport for NSW\footnote{\url{https://maps.transport.nsw.gov.au/egeomaps/traffic-volumes/index.html}}. The original PeMS collection covers nine districts (D03--D08 and D10--D12) through March 2024, and the EvoXXLTraffic extension described below adds records through 2025. PeMS supplies district-level daily files at 5-minute resolution. The inherited NSW collection was downloaded from its traffic-volume interface and organized at hourly or coarser resolution. Collection endpoints describe the archive, not continuous valid observations at every sensor.}

\begin{table}[!ht]
\centering
\color{blue}
\small
\caption{\revisiontext{Representative dataset subsets. Fixed-sensor counts and intervals follow the cited releases, cross-checked against the LargeST comparison~\cite{liu2024largest}. All rows use 5-minute intervals. ``Slots'' counts time indices, not sliding windows or valid sensor observations. EvoXXLTraffic coverage varies by district, and its 31-day benchmark windows are specified separately. The PEMS03, PEMS04, PEMS07, and PEMS08 rows are the published subsets of those names, not the EvoXXLTraffic districts.}}
\label{tab:data1}
\setlength{\tabcolsep}{4pt}
\begin{tabular}{llrl}
\toprule
Dataset & Source & Nodes & Archive interval (slots) \\
\midrule
METR-LA & DCRNN~\cite{li2018diffusion} & 207 & Mar--Jun 2012 (34,272) \\
PEMS-BAY & DCRNN~\cite{li2018diffusion} & 325 & Jan--Jun 2017 (52,116) \\
PEMSD7(M) & STGCN~\cite{yu2018spatio} & 228 & May--Jun 2012 (12,672)$^a$ \\
PEMSD7(L) & STGCN~\cite{yu2018spatio} & 1,026 & May--Jun 2012 (12,672)$^a$ \\
PEMS03 & STSGCN~\cite{song2020spatial} & 358 & Sep--Nov 2018 (26,208) \\
PEMS04 & ASTGCN/STSGCN~\cite{guo2019attention,song2020spatial} & 307 & Jan--Feb 2018 (16,992) \\
PEMS07 & STSGCN~\cite{song2020spatial} & 883 & May 1--Aug 6, 2017 (28,224) \\
PEMS08 & ASTGCN/STSGCN~\cite{guo2019attention,song2020spatial} & 170 & Jul--Aug 2016 (17,856) \\
LargeST-CA & LargeST~\cite{liu2024largest} & 8,600 & 2017--2021 (525,888) \\
LargeST-GLA & LargeST~\cite{liu2024largest} & 3,834 & 2017--2021 (525,888) \\
LargeST-GBA & LargeST~\cite{liu2024largest} & 2,352 & 2017--2021 (525,888) \\
LargeST-SD & LargeST~\cite{liu2024largest} & 716 & 2017--2021 (525,888) \\
\midrule
EvoXXLTraffic & This article & Yearly & Up to 1999--2025, nine districts$^b$ \\
\bottomrule
\end{tabular}
\par\smallskip\footnotesize
$^a$ Weekday subset. The slot count does not cover every calendar day.
$^b$ 213 district-year catalogs, with separate sizes and coverage in Table~\ref{tab:evo_per_district}. The 2025 district catalogs contain 572--4,888 sensors, totaling 19,672. This total is not one jointly trained graph.
\end{table}

\subsubsection{Fixed-Sensor Preprocessing}
\revisiontext{XXLTraffic selects long-lived sensor identities for comparisons across distant periods and derives hourly and daily aggregates~\cite{yin2025xxltraffic}. Identity membership does not establish complete observation coverage throughout the archive. The original processing code and download information are at \url{https://github.com/cruiseresearchgroup/XXLTraffic}. Section~\ref{sec:license} describes the corrected EvoXXLTraffic code, released data products, and their licenses.}

\revisiontext{A fixed sensor set makes non-contiguous windows spatially comparable but omits locations outside the selected intersection. EvoXXLTraffic instead preserves a separate catalog for each year, enabling the benchmark to state which sensors are retained, newly entering, returning, or absent.}

\subsection{EvoXXLTraffic: Sensor-Evolving Yearly Subsets}
\label{subsec:evoxxltraffic}

\revisiontext{The fixed-sensor subsets in XXLTraffic are designed for controlled long-horizon and gap-based forecasting, where the same set of sensors must be available across distant observation and prediction periods. However, this design hides an operationally important property of real traffic sensing systems: the sensor network itself evolves over time. The recorded PeMS catalogs change between years. These changes can reflect detector deployment, recording availability, or catalog selection, and the catalogs do not identify the cause of an individual change. To expose this evolution, we construct EvoXXLTraffic, a sensor-evolving reorganization of XXLTraffic.}

\revisiontext{EvoXXLTraffic follows the same PeMS data collection procedure described in Section~\ref{sec:data_collection}, but extends the raw PeMS crawl to the end of 2025. Instead of selecting only long-lived sensors, EvoXXLTraffic uses a separate sensor catalog for each year. For each PeMS district and each year, we construct a traffic-flow matrix, an ordered sensor-ID list, and a graph snapshot. This design supports studying forecasting under sensor-network evolution, including adaptation to sensors newly entering the yearly catalog. Identity alignment follows recorded station IDs. Physical relocation or reuse of an ID has not been independently verified over the complete archive.}

\subsubsection{Traffic Flow Data Preprocessing}
\label{subsubsec:evo_flow_preprocessing}

\revisiontext{For each PeMS district $d \in \{03,04,05,06,07,08,10,11,12\}$ and each calendar year $y$, we collect the available 5-minute daily records. The preprocessing defines the yearly sensor catalog from the sorted station IDs in one anchor day's records: the first available day for the district's first year, and January 1 for subsequent years. Daily records are then restricted to this catalog. It is not the union of every sensor appearing during the year, and later arrivals may only enter a subsequent year's catalog. Unlike the fixed-sensor XXLTraffic subsets, the catalog is allowed to change between years.}

\revisiontext{After identifying the yearly sensor catalog, we align the daily files to a regular 5-minute grid with persistent sensor IDs. The historical preprocessing stores a filled matrix, using forward fill, backward fill, and a zero fallback. This representation alone cannot distinguish unavailable records from valid zero traffic, and backward filling can use future information. The corrected benchmark therefore rebuilds its selected windows from the daily records and stores explicit masks alongside the flow values. This procedure is used for the 13-method comparisons on PEMS04 (five seeds) and PEMS05 (three seeds) and for the seven-method comparisons on the other seven districts (three seeds) described in Section~\ref{sec:result_evo}. Its target policy accepts a value only when the station record is present and its published Total Flow is finite. Valid zeros are retained, while absent or nonfinite values are excluded from training loss and evaluation. Published finite values may include upstream imputation, so this policy does not assert that every accepted value is a direct detector measurement. Original quality fields are retained for sensitivity analysis. Such quality-filtered sensitivity has not yet been evaluated in the reported comparison.}

\revisiontext{Each historical yearly matrix has shape $T_y\times N_y$, with a separate ordered sensor list. The corrected benchmark files also record target validity, input availability, timestamps, split boundaries, and preprocessing provenance. Inputs use forward-only filling. Before any value is available for a sensor, the training mean is used as a placeholder. The scalar mean and standard deviation are estimated only from accepted values at unique training slots. Target placeholders never contribute to the masked loss or metrics. The current model inputs contain filled flow histories, and the recorded input-availability mask is not supplied as an additional learned feature channel. Catalog size counts stored columns and must not be interpreted as the number of observed sensors at every time slot.}

\revisiontext{The corrected benchmark splits are generated from these explicitly masked windows according to Section~\ref{sec:benchmark_protocols}. The historical filled matrices remain available with the dataset, but are not the target-validity source for the benchmark runs.}

\subsubsection{Yearly Graph Preprocessing}
\label{subsubsec:evo_graph_preprocessing}

\revisiontext{EvoXXLTraffic also provides a graph snapshot for each district-year pair. For year $y$, the graph is constructed over the stored yearly sensor catalog. We use the PeMS sensor metadata as the source of geographic coordinates for these station IDs. The graph size therefore follows the catalog size, rather than remaining fixed across the full dataset span.}

\revisiontext{For each pair of catalog sensors, we compute their Haversine distance using latitude and longitude. We then convert pairwise distances into edge weights with a thresholded Gaussian kernel. Edges with weights below the sparsification threshold are removed. Following the convention used by existing streaming traffic benchmarks, we set the threshold to $\epsilon=0.1$. Self-loops are removed during preprocessing, allowing each downstream graph model to add self-loops according to its own implementation convention. If a sensor has missing geographic coordinates, the stored graphs substitute the mean coordinates of sensors with known locations. This fallback can introduce artificial proximity and is retained as an explicit limitation of the main comparison.}

\revisiontext{For reproducibility, the kernel of the stored graphs is $w_{ij}=\exp(-d_{ij}^{2}/\sigma_y^{2})$, with entries below 0.1 removed and diagonal entries set to zero. Distances $d_{ij}$ are Haversine distances in kilometers with radius 6,371 km, and $\sigma_y$ is the population standard deviation of positive pairwise distances in that year's construction coordinates. Thus a fixed weight threshold does not imply a fixed distance radius: the effective radius is $\sigma_y\sqrt{-\log(0.1)}$. Reconstructing distances from the thresholded adjacency alone would lose information. An unthresholded representation instead requires ordered sensor IDs, source coordinates and their dates, and explicit missing-coordinate flags. Missing distances then stay unavailable rather than being given a substitute location, unlike the centroid-filled graph used in the main comparison.}

\revisiontext{This graph encodes geographic proximity, not road links, directionality, or travel time. Adding or removing sensors can change neighborhoods even if the physical road network is unchanged. The fixed threshold is a benchmark construction choice, not evidence that one graph rule is optimal across years. The present experiments do not establish robustness to alternative thresholds or unthresholded distances.}

\begin{table}[t]
\centering
\color{blue}
\small
\caption{\revisiontext{Missing-coordinate fallback in the 213 stored yearly graph snapshots. A sensor-year counts one sensor in one yearly catalog, so repeated sensors contribute once per year. These are stored-data statistics, including PEMS04/2001, rather than the benchmark-only population.}}
\label{tab:coordinate_fallback}
\begin{tabular}{lrrr}
\toprule
District & Sensor-years & Fallback count & Fallback (\%) \\
\midrule
PEMS03 & 24,359 & 9 & 0.037 \\
PEMS04 & 63,882 & 618 & 0.967 \\
PEMS05 & 4,707 & 0 & 0.000 \\
PEMS06 & 8,024 & 0 & 0.000 \\
PEMS07 & 107,243 & 115 & 0.107 \\
PEMS08 & 33,215 & 0 & 0.000 \\
PEMS10 & 16,185 & 0 & 0.000 \\
PEMS11 & 30,944 & 32 & 0.103 \\
PEMS12 & 54,517 & 0 & 0.000 \\
\bottomrule
\end{tabular}
\end{table}

\revisiontext{Table~\ref{tab:coordinate_fallback} quantifies the missing-coordinate fallback over all 213 stored district-year snapshots. The largest pooled fraction is PEMS04's 618/63,882 sensor-years (0.967\%). This is not the percentage of distinct sensors or a bound on every individual year. A small fallback fraction also does not imply a small prediction effect. In PEMS04, 19,450/63,882 sensor-years use coordinate snapshots later than the catalog anchor, which is a separate issue from missing coordinates: the stored PEMS04 graphs are built from four metadata snapshots, whereas those of the other districts draw on 77 to 1,474 dated snapshots each. The released coordinate files record the latitude, longitude, and source snapshot of every sensor-year except the excluded PEMS04/2001 catalog, and they reproduce the other 212 stored graphs entry for entry. These counts motivate the explicitly retrospective interpretation below.}

\revisiontext{Each corrected yearly benchmark file stores explicit sensor IDs aligned to its flow columns and graph, and the benchmark runs verify this ordering and the integrity of the data and graphs. The main comparison uses retrospective geographic-proximity graphs and the anchor-day catalogs. Some coordinate metadata snapshots postdate the anchor day, and an anchor day's full station list need not be available at its beginning. These graphs and catalogs were therefore not necessarily available at each forecast date. Section~\ref{subsec:coordinate_sensitivity} reports a three-seed sensitivity analysis using coordinate snapshots dated no later than each anchor and isolating sensors without eligible coordinates. It holds the retrospective catalog fixed, so strictly prospective catalog availability remains unverified.}

\begin{figure*}[!h]
    \centering
    \includegraphics[width=0.99\textwidth]{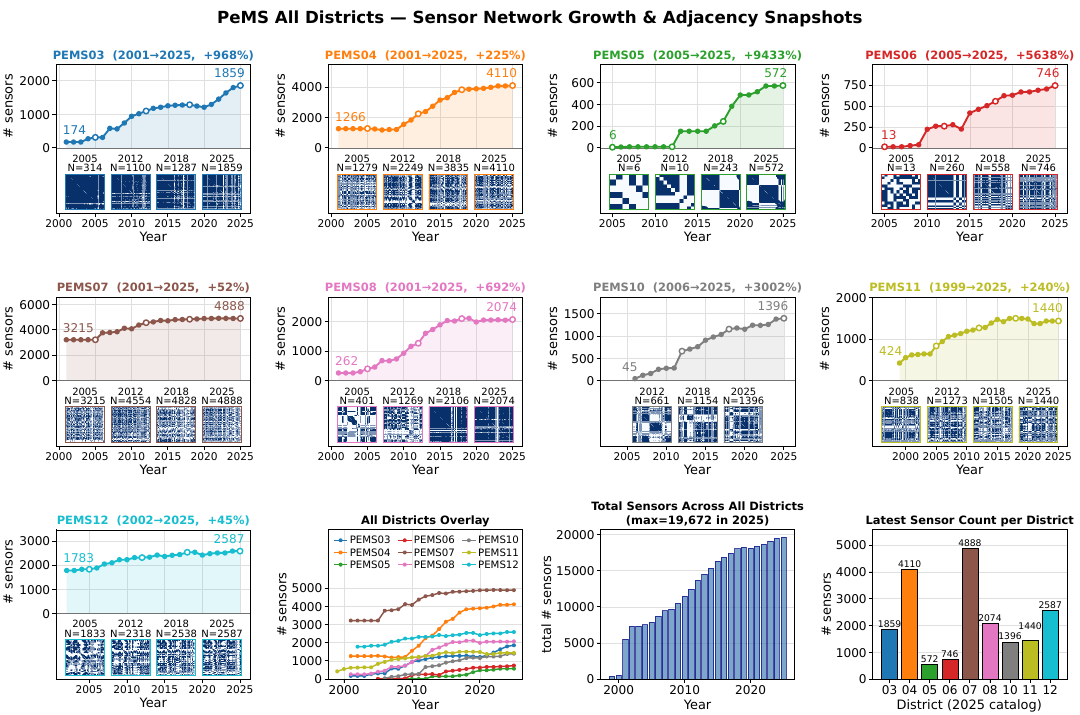}
    \caption{\revisiontext{Sensor-network growth and adjacency snapshots across PeMS districts. Each of the nine district panels shows the stored yearly catalog size of one district, together with representative yearly adjacency snapshots. Catalog size counts matrix columns, not sensors with valid observations at every slot. The PEMS04 visualization retains 2001, but its benchmark starts in 2002.}}
    \Description{Per-district yearly catalog sizes and representative geographic adjacency snapshots. PEMS04 stored coverage includes 2001.}
\label{fig:evo_growth_snapshots}
\end{figure*}

\subsubsection{Sensor Growth and Dynamic Graph Statistics}
\label{subsubsec:evo_growth_statistics}

\revisiontext{The yearly preprocessing reveals substantial differences in how PeMS districts evolve. As shown in Figure~\ref{fig:evo_growth_snapshots}, some districts experience extreme sensor expansion, while others are already large in the early years and grow more moderately. For example, PEMS05 and PEMS06 start from very small early-year graphs and expand by orders of magnitude, motivating evaluation of adaptation as the sensor catalog changes. In contrast, PEMS07 and PEMS12 have larger initial catalogs and smaller proportional changes. This heterogeneity could support future controlled comparisons of graph size and catalog turnover, but the present node counts alone do not separate their effects. The detailed evolution and growth statistics for each individual district are also provided in Table~\ref{tab:evo_per_district}.}

\revisiontext{Here $N_{\mathrm{first}}$ and $N_{\mathrm{last}}$ count the columns of the first and last stored yearly matrices, and \mbox{$100(N_{\mathrm{last}}/N_{\mathrm{first}}-1)\%$} is the growth. The counts are aligned to the saved yearly catalogs used for the growth visualization. Large growth ratios in initially small catalogs describe archive coverage and detector deployment, not a comparable multiplication of physical roads. Recent-period predictive behavior is examined separately in Section~\ref{subsec:evo_variation}. Calendar period alone does not identify a stable-road or stable-sensor regime.}

\revisiontext{The adjacency snapshots further show that EvoXXLTraffic has changing graph structures as well as changing node counts. The spatial graph is reconstructed over each yearly sensor catalog, producing year-specific graph topologies and node neighborhoods. Models evaluated on EvoXXLTraffic therefore encounter both temporal distribution shift and changes in the provided graph.}

\begin{table}[!ht]
\centering
\color{blue}
\small
\caption{\revisiontext{EvoXXLTraffic per-district processing summary. Years describe stored coverage, and node counts are columns in stored yearly matrices. Benchmark eligibility is specified in Section~\ref{subsec:evo_streaming_protocol}.}}
\label{tab:evo_per_district}
\setlength{\tabcolsep}{7pt}
\begin{tabular}{lcrrrr}
\toprule
District & Stored years & \# years & $N_{\mathrm{first}}$ & $N_{\mathrm{last}}$ & Growth \\
\midrule
PEMS03 & 2001--2025 & 25 & 174 & 1,859 & $+968\%$ \\
PEMS04 & 2001--2025 & 25 & 1,266 & 4,110 & $+225\%$ \\
PEMS05 & 2005--2025 & 21 & 6 & 572 & $+9{,}433\%$ \\
PEMS06 & 2005--2025 & 21 & 13 & 746 & $+5{,}638\%$ \\
PEMS07 & 2001--2025 & 25 & 3,215 & 4,888 & $+52\%$ \\
PEMS08 & 2001--2025 & 25 & 262 & 2,074 & $+692\%$ \\
PEMS10 & 2006--2025 & 20 & 45 & 1,396 & $+3{,}002\%$ \\
PEMS11 & 1999--2025 & 27 & 424 & 1,440 & $+240\%$ \\
PEMS12 & 2002--2025 & 24 & 1,783 & 2,587 & $+45\%$ \\
\bottomrule
\end{tabular}
\end{table}

\revisiontext{Figure~\ref{fig:evo_growth_snapshots} also summarizes growth across districts. The overlay compares district trajectories, the total-count view sums the available district catalogs in each year, and the final panel shows the 2025 counts. District coverage begins in different years, so a change in the total can also reflect an additional district entering the archive. Together, these statistics motivate the sensor-evolving streaming protocol introduced in Section~\ref{sec:benchmark_protocols}: a realistic long-term traffic benchmark should evaluate forecasting across long temporal spans and also expose the changing sensor network on which the forecasts are made.}


\subsection{Data Overview}
\label{subsec:data_overview}

\begin{figure}[!htbp]
  \centering
  \begin{minipage}[b]{0.39\linewidth}
    \centering
    \includegraphics[width=\linewidth]{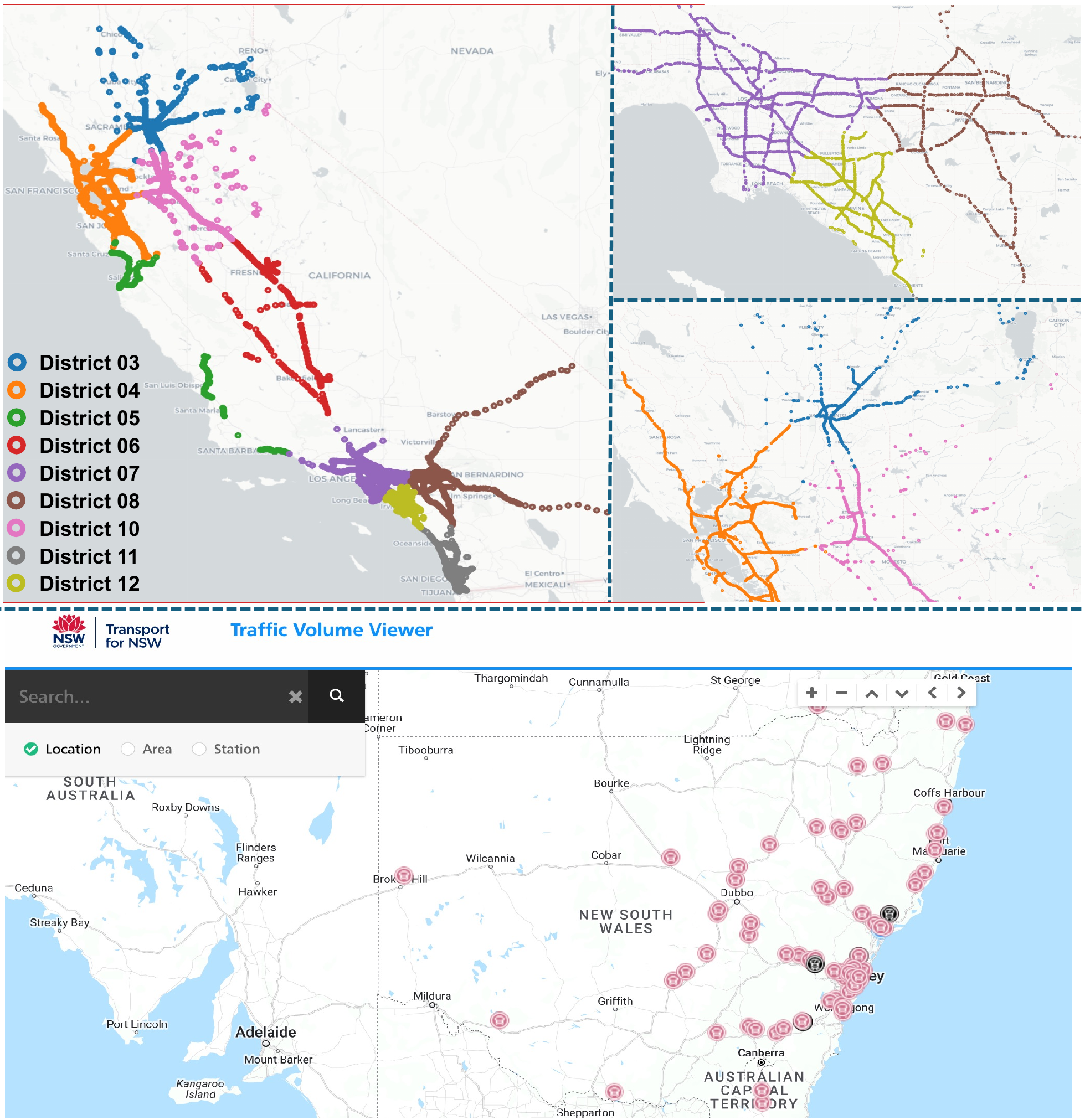}
    \vspace{-5pt}
    \subcaption{Global and local overview of XXLTraffic.}
    \label{fig:fig3}
  \end{minipage}
  \hfill
  \begin{minipage}[b]{0.60\linewidth}
    \centering
    \begin{subfigure}{\linewidth}
      \centering
      \includegraphics[width=\linewidth]{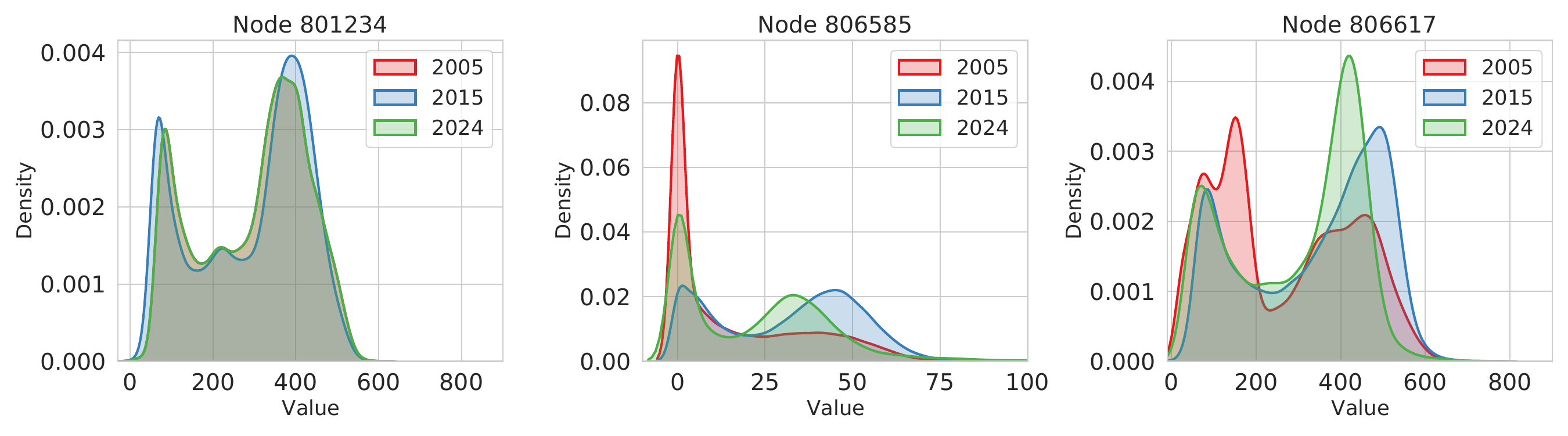}
      \vspace{-5pt}
      \caption{Distribution of District 8 in PeMS.}
      \label{fig41}
    \end{subfigure}
    
    \vspace{8pt}
    
    \begin{subfigure}{\linewidth}
      \centering
      \includegraphics[width=\linewidth]{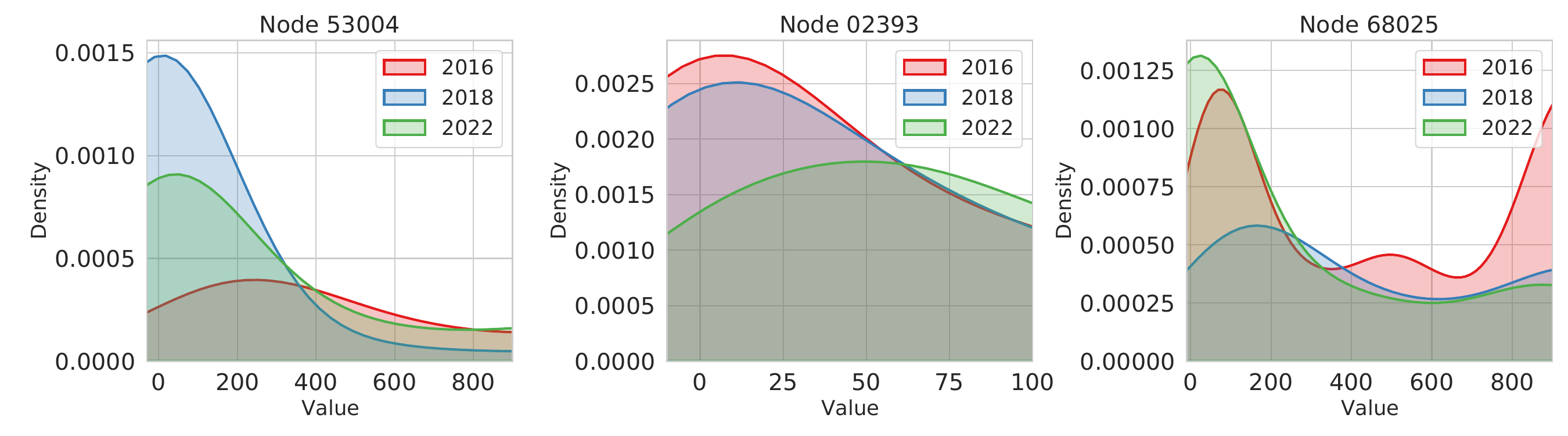}
      \vspace{-5pt}
      \caption{Distribution of Transport for NSW stations.}
      \label{fig42}
    \end{subfigure}
  \end{minipage}

  \vspace{5pt}
  \caption{\revisiontext{Historical XXLTraffic illustrations. (a) California and NSW sensor layouts. (b) Selected PeMS District 8 station distributions in 2005, 2015, and 2024. (c) Selected NSW station distributions in 2016, 2018, and 2022. These examples are inherited visualizations, not quality-filtered analyses of the corrected benchmark targets.}}
  \Description{Regional sensor layouts and historical traffic status distributions for PeMS District 8 and Transport for NSW.}
\label{fig:fig4} 
\end{figure}

\revisiontext{Figure~\ref{fig:fig4}(a) illustrates the regional coverage of the inherited California and NSW collections. The map locates recorded sensors but does not establish road-link connectivity or measurement completeness. The density plots in panels (b,c) illustrate differing historical value distributions at selected station IDs. They are descriptive examples, not district-wide tests of distribution shift or evidence that urban construction caused a particular change. Their measurement and missing-value conventions are inherited from the original visualization, separate from the corrected benchmark masks.}

\subsection{Dataset and Code Licensing}
\label{sec:license}
\revisiontext{EvoXXLTraffic is released with explicit licenses. The derived data products are released under CC BY-NC 4.0, consistent with XXLTraffic. They comprise the yearly catalogs, the corrected observation-aware benchmark files, the graph snapshots, the sensor coordinates, the run configurations, and the result tables. Author-written code is released under the MIT License. Adapted third-party baseline code keeps the license of its original repository, and the PeMS terms continue to apply to the upstream records. A notice in the code repository records these distinctions.}

\revisiontext{The code and results of this article are available at \url{https://github.com/cruiseresearchgroup/TSAS26-EvoXXLTraffic}. The repository contains:
\begin{itemize}
\item the frozen code of all 13 adaptations
\item the run configurations, with the checksums of their data and graphs
\item the per-year results of all runs, from which every reported mean and standard deviation can be recomputed
\item the training-free references and the sensor-coordinate files
\item the code and annual results of the fixed- versus evolving-catalog control (Section~\ref{subsec:fixed_vs_evolving})
\end{itemize}
The Hugging Face dataset at \url{https://huggingface.co/datasets/yind7/EvoXXLTraffic} contains the corrected benchmark files, the 3,523 dated metadata snapshots, and the coordinates. The benchmark files hold explicit target and input masks, and their checksums equal those recorded in the run configurations. The dataset also provides the full-year flow matrices and graphs of all 213 district-years. These matrices are stored filled, as described in Section~\ref{subsubsec:evo_flow_preprocessing}, and must not replace the masked benchmark files for evaluation.}

\section{Benchmark Protocols}
\label{sec:benchmark_protocols}
\begin{figure}[!h]
\centering
\includegraphics[width=0.74\linewidth]{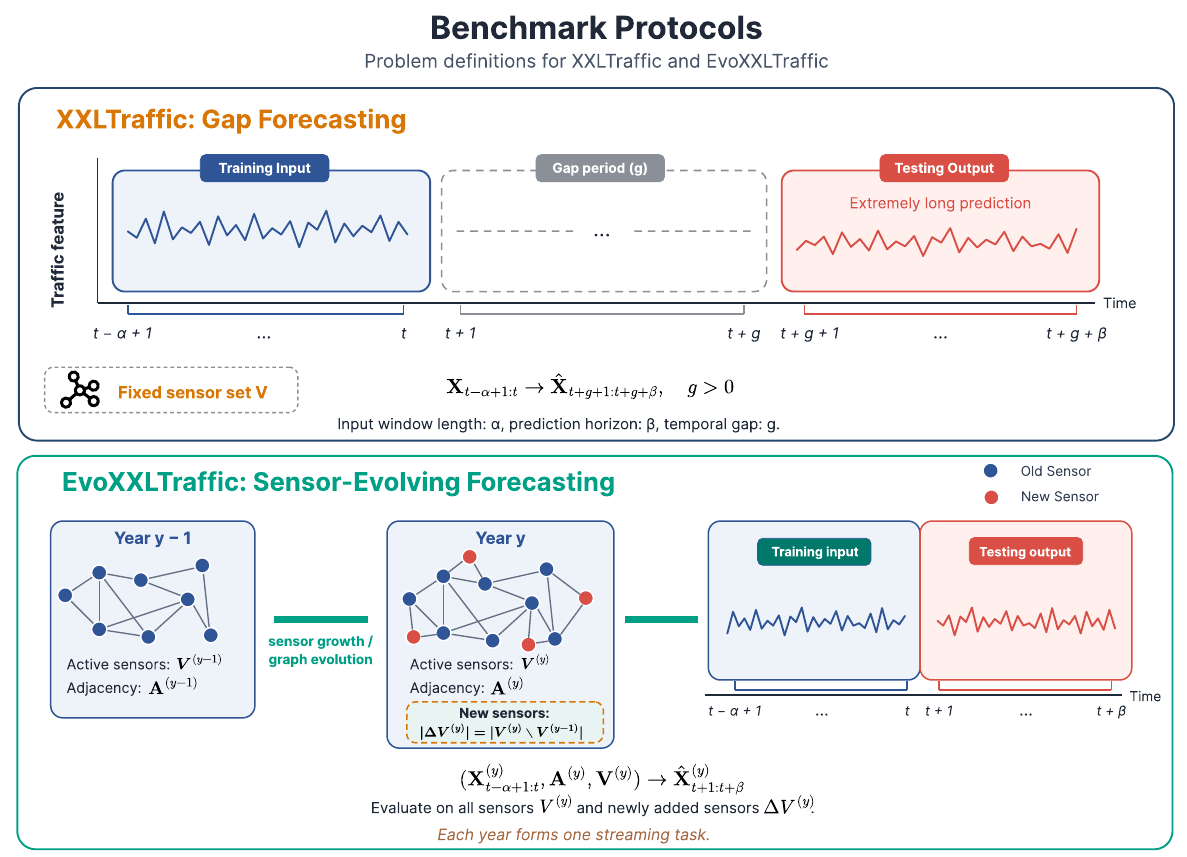}
\caption{\revisiontext{Conceptual forecasting tasks. The left/right windows denote model inputs and prediction targets, rather than the complete training/test splits. In the EvoXXLTraffic schematic, ``active'' denotes catalog membership, and new sensors may have current-year training observations. The actual 31-day splits and label-update policy are specified in Section~\ref{subsec:evo_streaming_protocol}.}}
\Description{Comparison of fixed-sensor temporal-gap and yearly sensor-evolving forecasting protocols.}
\label{fig:exp_proto}
\end{figure}

\revisiontext{Figure~\ref{fig:exp_proto} contrasts gap and sensor-evolving forecasting. These two settings and hourly/daily aggregation ask complementary questions. Fixed-sensor gap forecasting studies temporal separation at matched sensor identities, while hourly/daily aggregation changes temporal resolution. Sensor-evolving forecasting changes the yearly sensor population and the associated graph. The first two are inherited settings, and the new journal benchmark concerns the third. Their metrics and evaluation populations are not pooled, and success in one setting does not establish success in the others.}

\subsection{Fixed-Sensor Extremely Long Forecasting Protocols}
\label{subsec:fixed_protocols}

\revisiontext{XXLTraffic uses an early-installed fixed sensor subset for gap forecasting and derives hourly and daily aggregations from this subset, as shown in Table~\ref{tab:fixed_sensor_protocols}. The historical gap experiments use gaps of 1, 1.5, and 2 years and a 6:2:2 chronological train/validation/test split, while the hourly/daily experiments use 7:1:2. In the gap experiments, the default 10\% sampling operates on windows in every split: training and validation use randomized subsets, while testing uses the first 10\% of test windows without shuffling. Thus it is not uniform sampling of all time slots or full-test evaluation. The nominal input length is 96 and prediction lengths are 96/192/336. The results of these settings are reported in the conference paper and are not reproduced here (Appendix~\ref{app:legacy_results}).}

\begin{table}[!h]
\centering
\small
\setlength{\tabcolsep}{4pt}
\renewcommand{\arraystretch}{0.95}
\caption{Fixed-sensor forecasting protocols retained from XXLTraffic.}
\label{tab:fixed_sensor_protocols}
\begin{tabular}{@{}lccc@{}}
\toprule
\textbf{Protocol} & \textbf{Sensor set} & \textbf{Temporal relation} & \textbf{Main purpose} \\
\midrule
Gap forecasting & Fixed long-lived sensors & Non-contiguous, $g>0$ & Long-term distribution shift \\
Hourly forecasting & Fixed long-lived sensors & Adjacent window & Coarse-scale long-term forecasting \\
Daily forecasting & Fixed long-lived sensors & Adjacent window & Long-range trend forecasting \\
\bottomrule
\end{tabular}
\end{table}

\subsection{Sensor-Evolving Streaming Forecasting Protocol}
\label{subsec:evo_streaming_protocol}

\revisiontext{EvoXXLTraffic evaluates forecasting under yearly sensor-network evolution. A yearly task uses the first 31 stored days of that year, rather than a full calendar year. The model receives the task's training split, retrospective adjacency $A_y$, and sensor-ID mapping, then forecasts the chronological test split. Table~\ref{tab:evo_streaming_protocol} specifies the corrected protocol used for the 13-method comparisons on PEMS04 (five seeds) and PEMS05 (three seeds) and for the seven-method comparisons on the other seven districts (three seeds).}

\noindent\textbf{Dataset coverage and benchmark eligibility.}
\revisiontext{The temporal coverage of the released data and visualizations is distinct from the years used for benchmarking. In particular, we retain the PEMS04 data and sensor-network visualization for 2001, but define its benchmark period as 2002--2025 for all methods and seeds. The stored 2001 matrix contains nonzero values later in the year, whereas the first 31 days used by the benchmark yield all-zero training, validation, and test targets. We therefore exclude this task from training and evaluation, and do not use checkpoints trained on it to initialize the benchmark. The other eight districts retain the starting years listed in Table~\ref{tab:evo_per_district}. An audit of their first two stored years found nonzero targets in each split. This check establishes that these tasks are not entirely zero, but does not establish complete observation coverage or distinguish missing-value fills from genuine zero traffic flow.}

\noindent\textbf{Metric and aggregation conventions.}
\revisiontext{For the corrected runs, each split contains complete 12-step input/12-step target windows, with no window crossing a split boundary. The 8,928 slots are split into 5,356 training, 1,786 validation, and 1,786 test slots, yielding 5,333/1,763/1,763 complete windows. All 1,763 test origins per year are evaluated chronologically. Past-value filling may carry observations from an earlier split, but never from a later time. Mean absolute error (MAE) and root mean squared error (RMSE) are computed on published Total Flow counts per station per 5-minute interval, including accepted zeros. Mean absolute percentage error (MAPE) is expressed as a percentage and excludes targets with absolute value at most $10^{-6}$. Its denominator is therefore reported separately in the released results. A horizon entry $h\in\{3,6,12\}$ pools the forecast--target pairs at leads $1,\ldots,h$. It is not the error at lead $h$ alone. Avg is the full 12-lead pool and consequently equals the entry for 12. Within a seed, yearly scores receive equal weight over years with valid targets, and empty cohorts are N/A, not zero error. We report the mean and sample standard deviation across seeds 47--51 for PEMS04 and across seeds 47--49 for the other districts. Seed variation is descriptive and is not a confidence interval or a test of statistical significance.}

\noindent\textbf{Coverage limitation.}
\revisiontext{PEMS04 begins with 1,266 catalog columns in 2002 and reaches 4,110 in 2025. Only 16.27\% of its possible test forecast--target pairs in 2002 pass the main validity mask. Thus the initial task is sparse in accepted targets despite its large catalog, so node count alone cannot explain model performance.}

\begin{table}[!h]
\centering
\small
\setlength{\tabcolsep}{5pt}
\renewcommand{\arraystretch}{0.95}
\caption{\revisiontext{Corrected sensor-evolving protocol.}}
\label{tab:evo_streaming_protocol}
\begin{tabular}{@{}ll@{}}
\toprule
\textbf{Component} & \textbf{Setting} \\
\midrule
Task unit & \revisiontext{One yearly 31-day task} \\
Sensor set & \revisiontext{Stored yearly sensor catalog} \\
Graph & \revisiontext{Retrospective geographic adjacency $A_y$} \\
Sensor mapping & \revisiontext{Retained, newly entering, returning, absent} \\
In-year window & \revisiontext{First 31 stored days, $31 \times 288 = 8{,}928$ slots} \\
Split & 60/20/20 chronological train/val/test \\
Input / output length & 12 / 12 steps \\
Normalization & \revisiontext{$z$-score from accepted training slots only} \\
Evaluation split & Chronological test split of each year \\
Metrics & MAE, RMSE, MAPE \\
Horizons & \revisiontext{Cumulative 1--3, 1--6, 1--12 (Avg = 1--12)} \\
Evaluation groups & \revisiontext{All, new to preceding year (incl. returning)} \\
\bottomrule
\end{tabular}
\end{table}

\revisiontext{Here ``online'' in Online-AN and Online-NN means sequential yearly fine-tuning on the current training split. These models do not update their parameters with test labels. ST-TTC alone performs supervised calibration during testing, after an earlier prediction's complete 12-step target window has arrived (60 minutes after its origin). All evaluation cohorts are scored on the same predictions. This feedback policy is part of the comparison, not an equal-information ablation of adaptation strategies.}

\section{Experiments}
\label{exp_section6}

\revisiontext{This section evaluates the sensor-evolving benchmark on the nine-district scope of Section~\ref{subsec:research_questions}. The main text reports compact comparisons and analyses addressing RQ1--RQ3, and Appendix~\ref{app:evo_details} contains the detailed horizon/metric tables. The inherited fixed-sensor gap/hourly/daily experiments are reported in the conference paper~\cite{yin2025xxltraffic}, and Appendix~\ref{app:legacy_results} explains why they are not reproduced. They are not evidence for the corrected EvoXXLTraffic comparisons or part of the new experimental contribution.}

\subsection{Baselines}

\label{subsec:baseline_scope}
\revisiontext{We evaluate 13 frozen baseline adaptations, grouped by their update policy. Table~\ref{tab:baseline_implementation} specifies the executed mechanisms and their remaining adaptations. The comparison concerns these configurations under the common data and evaluation protocol and does not reproduce each original paper's tuning or experimental setting.}

\noindent\revisiontext{\textbf{Static adaptations:} DCRNN~\cite{li2018diffusion}, ASTGCN~\cite{guo2019attention}, and TGCN~\cite{zhao2019t}, each initialized anew for every yearly task.}

\noindent\revisiontext{\textbf{Training schemes:} Pretrain fits the first task only, and Retrain initializes anew each year. Online-NN fine-tunes using new-to-previous-year loss targets, whereas Online-AN fine-tunes using all current training targets. These four share a common graph/feature-convolution predictor.}

\noindent\revisiontext{\textbf{Continual adaptations:} TrafficStream~\cite{ijcai2021p0498} uses node selection, replay, and regularization. PECPM~\cite{wang2023pattern} uses a learned memory predictor with training-derived pattern supervision and node selection. STKEC~\cite{wang2023knowledge} combines its memory predictor with training-only influence selection using a damped generalized Gauss--Newton (GGN) approximation. EAC~\cite{chen2025expand} freezes its backbone after the first year and updates sensor-aligned low-rank parameters.}

\noindent\revisiontext{\textbf{Retrieval/calibration adaptations:} STRAP~\cite{zhang2025strap} retrieves from spatial, temporal, and spatio-temporal pattern libraries carried across yearly tasks. ST-TTC~\cite{chen2025learning} adds a spectral calibrator with delayed test-label updates to a newly fitted yearly predictor.}

\begin{table}[ht]
\centering
\color{blue}
\small
\caption{\revisiontext{Executed baseline mechanisms and benchmark adaptations for all 13 methods. ASTGCN uses the recent-only branch, and PECPM and STKEC retain the stated training-only adaptations. These rows do not imply reproduction of the original papers' experimental settings or optimized rankings.}}
\label{tab:baseline_implementation}
\begin{tabular}{@{}lp{0.76\linewidth}@{}}
\toprule
Method & Implemented computation and scope \\
\midrule
DCRNN & A two-layer diffusion GRU encodes all 12 input times and autoregressively decodes 12 outputs. Bidirectional diffusion uses separate normalized supports and order-2 Chebyshev recurrences. Scheduled sampling uses observed training labels, and decoder inputs and outputs use the training scaler. \\
\addlinespace
ASTGCN & The author-released recent-only architecture uses two blocks with sample-specific temporal/spatial attention, order-3 graph Chebyshev convolution, temporal convolution, residuals, and layer normalization. Graph polynomials use matrix recurrence. Daily/weekly branches are not used. \\
\addlinespace
TGCN & Self-loop, symmetric-normalized graph convolution enters the recurrent gates at every input time. Gates split the feature axis per node, and a shared linear head maps the final state to 12 outputs. \\
\addlinespace
Pretrain & The common graph/feature-convolution predictor shared by the four training schemes, fitted on the first benchmark year only (PEMS04: 2002) and applied unchanged afterwards. \\
\addlinespace
Retrain & The common predictor, initialized anew and fitted on every yearly task. \\
\addlinespace
Online-NN & The common predictor, fine-tuned each year on new-to-previous-year loss targets with two-hop context. Fitting is skipped when that cohort is empty. \\
\addlinespace
Online-AN & The common predictor, fine-tuned each year on all current training targets. \\
\addlinespace
TrafficStream & The source predictor with node selection from training information, replay, and regularization. Its state carries across yearly tasks. \\
\addlinespace
PECPM & The author-released predictor uses trainable $64\times12$ memory and learned retrieval. Unavailable external pattern labels are reconstructed from training-prefix clustering and cosine matching. Stable-ID node selection, training-only drift/replay, and masked Fisher regularization adapt it to this benchmark. \\
\addlinespace
STRAP & The author-released backbone and spatial, temporal, and spatio-temporal libraries use aligned historical keys and backbone-derived values, topology features, and fusion. Libraries are built only from training origins and carried in checkpoints. \\
\addlinespace
STKEC & The memory predictor is retained. Influence selection uses disjoint training query/pool windows and a damped generalized Gauss--Newton (GGN) inverse, replacing the source detector's test-dependent queries. This is a training-only influence approximation. \\
\addlinespace
EAC & A station-indexed low-rank offset $UV$ modifies inputs, and the backbone freezes after year one. Stable-ID row alignment extends the authors' growing-node policy. \\
\addlinespace
ST-TTC & A predictor fitted anew each year plus a spectral calibrator that is updated during testing only after the complete 12-step target window of an earlier prediction is available. \\
\bottomrule
\end{tabular}
\end{table}

\revisiontext{We checked six implementations (DCRNN, ASTGCN, TGCN, PECPM, STRAP, and STKEC) against their original sources and corrected the omissions found before the runs reported here. Their 48 runs comprise 1,098 method--seed--year tasks. With the other 206 main and coordinate-policy runs and the 24 fixed- versus evolving-catalog control runs, the total is 278. We verified each mechanism, chronological two-year pilots, full-graph capacity, and checkpoint continuity before the formal runs. The adaptations in Table~\ref{tab:baseline_implementation} remain part of the benchmark definition. PECPM's reconstructed training-pattern supervision and STKEC's training-only GGN selector are not claimed to be identical to the original training pipelines.}

\subsection{Implementation Details}
\label{sec:exp_details}

\noindent\textbf{Datasets used.}
\revisiontext{The dataset contains all nine districts in Table~\ref{tab:evo_per_district}. The corrected main comparison covers 13 methods on PEMS04 and PEMS05. Pretrain, Retrain, Online-NN, Online-AN, TrafficStream, EAC, and ST-TTC are also evaluated on PEMS03, PEMS06, PEMS07, PEMS08, PEMS10, PEMS11, and PEMS12. Each district has separate training and test data, and no cross-district transfer is performed.}

\noindent\textbf{Sensor-evolving streaming setup.}
\revisiontext{The main PEMS04 runs use seeds 47--51. All other runs, including the PEMS04 coordinate-policy runs, use seeds 47--49. All use input/output lengths 12/12 and at most 100 training epochs per year. Training minimizes masked MSE plus each method's auxiliary terms using AdamW with weight decay 0.01 and no learning-rate scheduler. Early stopping selects the checkpoint with the lowest full-graph validation MAE over all accepted targets and all 12 leads, with a patience of five consecutive epochs without strict improvement. The selected model and auxiliary state are restored before testing. Optimizer state is initialized afresh in every yearly fitting stage. Continual methods load the immediately preceding year's checkpoint with matching method, seed, data, graph, and protocol provenance. No PEMS04 checkpoint from the excluded 2001 task is used.}

\revisiontext{The per-year budget means an epoch cap of 100, with the early-stopping rule above, rather than fixed wall time, FLOPs, or optimizer steps. An epoch traverses the method's eligible training windows and target population. Node selection, skipped fitting, auxiliary updates, and batch size can change the amount of computation. The recorded epochs, optimizer steps, and complete-run timing make these differences explicit.}

\revisiontext{Learning rates are 0.01 for DCRNN, PECPM, Pretrain, Online-NN, TrafficStream, and STKEC, and 0.03 otherwise. DCRNN clips the gradient norm at 5 and uses inverse-sigmoid scheduled sampling with constant 1,000, restricted to observed training targets. Batch size is 64 for EAC, TrafficStream, and STKEC, and 128 otherwise. These settings are fixed across seeds. Where memory requires smaller microbatches, losses are accumulated with observation-count weighting so that each update matches the stated batch size. On PEMS04, Pretrain fits only 2002. Retrain, DCRNN, ASTGCN, TGCN, and ST-TTC initialize anew each year. Online-AN fine-tunes on all current sensors. After the first year, Online-NN uses new-to-previous-year loss targets with two-hop context and skips fitting if that cohort is empty. EAC freezes its backbone after the first year. TrafficStream, PECPM, and STKEC select nodes using training information. PECPM and STRAP retain their pattern state. ST-TTC updates calibration only after the complete target window for an earlier prediction is available.}

\revisiontext{PECPM uses a 64-pattern memory, top-5 training-derived matching targets, pattern-loss weight 5, and elastic weight consolidation (EWC) weight 0.001. STRAP builds libraries from the first 128 chronological training origins, with history ratio 0.3, history limit 10,000, 16 retrieved patterns, and fusion weight 0.7. STKEC uses the last 16 nonoverlapping training query windows and a disjoint earlier pool, GGN damping 0.1, and up to 80 conjugate-gradient iterations, and failed solves are rejected. These settings, the data, and the graphs were frozen before the formal runs.}

\revisiontext{Architecture, auxiliary updates, learning rate, and batch size differ between methods, so this is not an isolation of streaming policy at equal capacity or computation. The released run records list each run's GPU and software environment. Elapsed times are not a common-hardware speed comparison: some runs used different GPU types or resumed from a checkpoint, which preserves model, optimizer, auxiliary, and random-number state within a yearly task. Every run keeps the frozen yearly sequence and evaluation population. No seed or checkpoint was chosen using test scores.}

\subsection{Results of Sensor-Evolving Streaming Forecasting}
\label{sec:result_evo}

\begin{figure}[t]
  \centering
  \includegraphics[width=\linewidth]{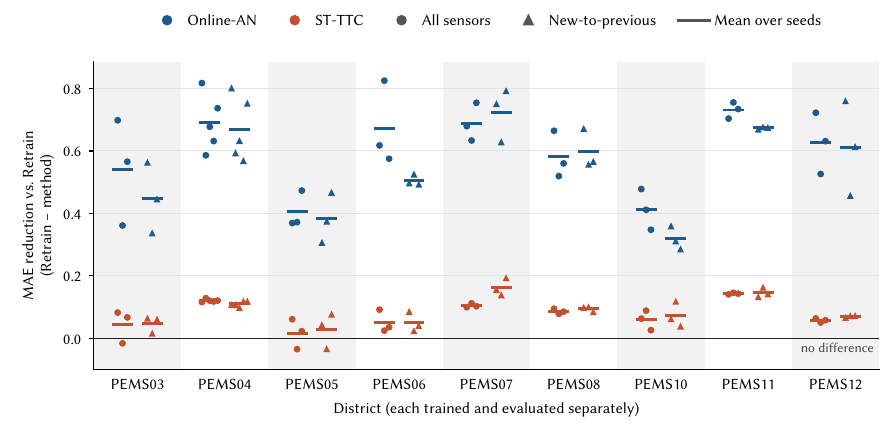}
  \Description{Per-seed MAE reductions of Online-AN and ST-TTC relative to Retrain in each of the nine districts, shown separately for all sensors and new-to-previous sensors, with dashes at the seed means.}
  \caption{\revisiontext{Paired MAE differences against per-year Retrain in all nine districts, each trained and evaluated independently, at the full 12-lead horizon. Each marker is one seed (47--49, or 47--51 on PEMS04) in one district and cohort: its height is Retrain~$-$~method for that seed (positive = lower MAE than Retrain), taken within the seed before averaging across seeds. Color marks the method and shape the cohort (circle for all sensors, triangle for new-to-previous), and a short dash marks the mean over seeds. Online-AN has a lower MAE than Retrain for every seed in all 18 district--cohort cells (mean reductions 0.32--0.73 MAE), whereas ST-TTC does so in 15 of 18 (mean reductions 0.02--0.16). The figure shows two of the seven methods evaluated in every district, not all 13 benchmark methods. Table~\ref{tab:tsas_three_seed_diagnostics} lists all seven in each district and cohort as mean $\pm$ sample SD. The SD measures dispersion and is not a confidence interval.}}
  \label{fig:district_paired_diffs}
\end{figure}

\revisiontext{The corrected comparison evaluates 13 methods over 24 yearly PEMS04 tasks (2002--2025), with five common seeds per method. All-sensor scores use every accepted target in the current sensor catalog. New-node scores use $\mathcal{V}_y\setminus\mathcal{V}_{y-1}$ (Equation~\ref{eq:new_sensors}), the new-to-previous-year cohort (shortened to new-to-previous in figures), including returning sensors. Every method except Pretrain has supervised current-year training first. A separate first-ever cohort excludes returning sensors. The released results keep these cohorts and their denominators separate. Table~\ref{tab:tsas_full_horizon_summary} summarizes the completed PEMS04 and PEMS05 comparisons at 12 leads, and Figure~\ref{fig:benchmark_leaderboard} ranks the 13 methods within each district and cohort. The detailed tables in Appendix~\ref{app:evo_details} retain all reported horizons and metrics. Figure~\ref{fig:district_paired_diffs} shows paired seed-level differences in all nine districts, and Table~\ref{tab:tsas_three_seed_diagnostics} lists the seven-method MAE values for every district. Unevaluated combinations are outside the reported scope, not failed runs.}

\revisiontext{The same protocol is applied to all 13 methods on PEMS05 (2005--2025) and to the seven methods on PEMS03 (2001--2025), PEMS06 (2005--2025), PEMS07 (2001--2025), PEMS08 (2001--2025), PEMS10 (2006--2025), PEMS11 (1999--2025), and PEMS12 (2002--2025), each with seeds 47--49. Each district uses its own training, validation, and test data, so these are separate within-district benchmarks, not transfer from a model trained on PEMS04. These 186 completed runs cover 4,326 method--seed--year tasks. They quantify seed variability across districts, but the seven-method districts do not establish rankings among all 13 baselines.}

\begin{table}[t]
\centering
\color{blue}
\small
\caption{\revisiontext{Full-12-lead MAE for all 13 methods. PEMS04 uses five seeds (47--51), PEMS05 three (47--49). Entries are mean$\pm$sample standard deviation after equal-year aggregation. ``New'' means new to the preceding year, including returning sensors, and every method except Pretrain has current-year supervised training. The all/new year counts are 24/20 for PEMS04 and 21/12 for PEMS05. Implementation adaptations in Table~\ref{tab:baseline_implementation} apply throughout. ASTGCN is recent-only. The two training-free references use the same accepted targets: each year's training mean predicted for every target, and the last input value repeated for all 12 leads. Bold marks the lowest unrounded mean among the 13 methods in each column, without a significance claim. See Section~\ref{sec:result_evo_allsensor} for TGCN.}}
\label{tab:tsas_full_horizon_summary}
\setlength{\tabcolsep}{5pt}
\begin{tabular}{lrrrr}
\toprule
 & \multicolumn{2}{c}{PEMS04} & \multicolumn{2}{c}{PEMS05} \\
\cmidrule(lr){2-3}\cmidrule(lr){4-5}
Method & All & New & All & New \\
\midrule
DCRNN & $24.460\pm0.393$ & $21.262\pm0.487$ & $12.484\pm0.295$ & $11.117\pm0.510$ \\
ASTGCN & $23.013\pm0.120$ & $19.908\pm0.132$ & $11.766\pm0.090$ & $10.377\pm0.151$ \\
TGCN & $130.504\pm2.156$ & $129.698\pm2.600$ & $50.522\pm0.320$ & $49.786\pm4.169$ \\
Pretrain & $42.103\pm1.609$ & $41.508\pm1.720$ & $27.436\pm0.171$ & $24.947\pm0.255$ \\
Retrain & $22.081\pm0.092$ & $18.688\pm0.070$ & $11.920\pm0.049$ & $10.367\pm0.049$ \\
Online-NN & $23.736\pm0.828$ & $17.797\pm0.161$ & $13.956\pm0.324$ & $10.044\pm0.055$ \\
Online-AN & $21.391\pm0.088$ & $18.018\pm0.069$ & $\mathbf{11.515\pm0.037}$ & $\mathbf{9.984\pm0.034}$ \\
TrafficStream & $22.013\pm0.366$ & $18.286\pm0.354$ & $13.705\pm0.050$ & $10.228\pm0.051$ \\
PECPM & $\mathbf{20.760\pm0.081}$ & $\mathbf{17.315\pm0.093}$ & $13.467\pm0.276$ & $10.002\pm0.054$ \\
STKEC & $22.516\pm0.192$ & $18.811\pm0.236$ & $12.106\pm0.056$ & $10.412\pm0.036$ \\
EAC & $24.505\pm2.596$ & $22.311\pm3.309$ & $14.062\pm0.593$ & $12.443\pm0.395$ \\
STRAP & $21.574\pm0.090$ & $18.156\pm0.090$ & $11.777\pm0.059$ & $10.202\pm0.052$ \\
ST-TTC & $21.960\pm0.094$ & $18.578\pm0.067$ & $11.903\pm0.038$ & $10.339\pm0.049$ \\
\midrule
\multicolumn{5}{l}{\textit{Training-free references (no seed variation)}} \\
Training mean & $134.806$ & $138.089$ & $61.318$ & $60.109$ \\
Last value & $27.749$ & $23.540$ & $15.270$ & $13.239$ \\
\bottomrule
\end{tabular}
\end{table}

\begin{figure}[t]
  \centering
  \includegraphics[width=\linewidth]{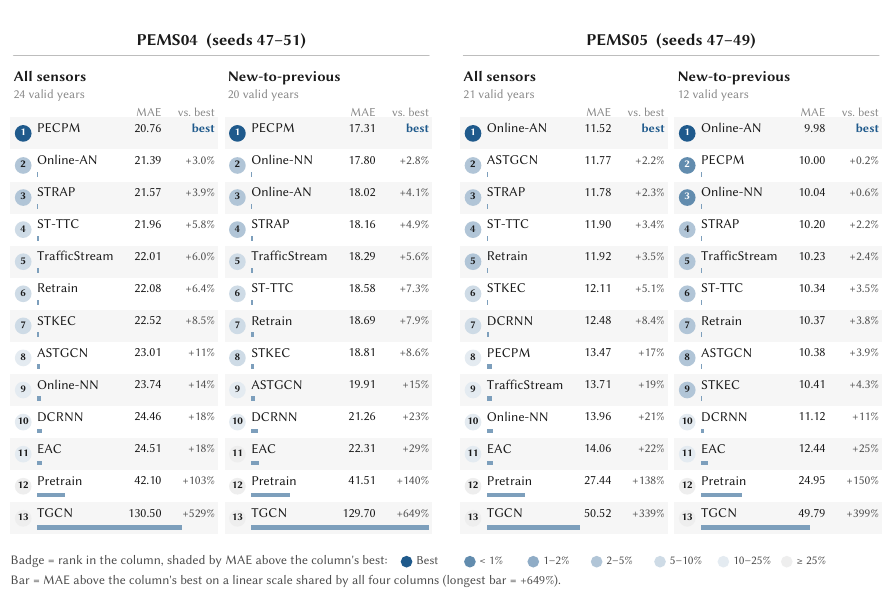}
  \Description{Four ranked lists, PEMS04 and PEMS05 for all and new-to-previous sensors, ordering the 13 methods by mean MAE with rank badges and bars for the gap to each list's best method.}
  \caption{\revisiontext{Full-horizon (leads 1--12 aggregated) MAE of the 13 methods on PEMS04 (seeds 47--51, with 24 valid all-sensor years and 20 valid new-to-previous years) and PEMS05 (seeds 47--49, with 21 and 12 valid years). Each column lists the methods from the lowest to the highest mean MAE, where every seed is first averaged with equal weight over its valid years and the mean is then taken across seeds. The badge gives the rank within the column and is shaded by how much higher the method's MAE is than the column's best. The bar and the percentage give the same gap on a linear scale shared by all four columns. The two districts are trained and evaluated separately, so columns are not transfer results, and ranks compare methods within a column, not cohorts. PECPM has the lowest MAE in both PEMS04 columns, and Online-AN in both PEMS05 columns. Online-NN ranks 9/2 on PEMS04 (all sensors / new-to-previous) and 10/3 on PEMS05, the largest cohort-dependent shift. Large errors (TGCN, last in every column and up to +649\% above the column's best) are the retained results of the frozen configurations (Section~\ref{sec:result_evo_allsensor}) and are drawn at full length. Method names denote the frozen implementations and adaptations in Table~\ref{tab:baseline_implementation} (ASTGCN is recent-only). Ranks summarize means and are not significance statements. The mean $\pm$ sample SD of every cell is in Table~\ref{tab:tsas_full_horizon_summary}.}}
  \label{fig:benchmark_leaderboard}
\end{figure}

\subsubsection{RQ1: All-Sensor Evaluation}
\label{sec:result_evo_allsensor}

\revisiontext{In the PEMS04 summary (Table~\ref{tab:tsas_full_horizon_summary} and Figure~\ref{fig:benchmark_leaderboard}, with detailed metrics in Table~\ref{tab:tsas_evo_part1}), PECPM has the lowest full-12-lead all-sensor MAE ($20.760\pm0.081$), followed by Online-AN ($21.391\pm0.088$) and STRAP ($21.574\pm0.090$). ST-TTC, TrafficStream, and Retrain achieve $21.960\pm0.094$, $22.013\pm0.366$, and $22.081\pm0.092$, respectively. PECPM also has the lowest 12-lead all-sensor RMSE and MAPE, although Online-AN leads all three metrics at horizon 3 (Table~\ref{tab:tsas_evo_part1}). Pretrain reaches $42.103\pm1.609$, above even the training-free last-value reference (27.749), while \mbox{Online-NN} reaches $23.736\pm0.828$.}

\revisiontext{TGCN has a high all-sensor MAE: $130.504\pm2.156$ on PEMS04 and $50.522\pm0.320$ on PEMS05. Its error hardly changes with the horizon (PEMS04 MAE 130.52, 130.51, and 130.50 at horizons 3, 6, and 12) or between cohorts, the pattern of an almost input-independent forecast. The training-free references in Table~\ref{tab:tsas_full_horizon_summary} make this concrete: predicting the training mean everywhere gives 134.806 on PEMS04 and 61.318 on PEMS05, whereas repeating the last input value gives 27.749 and 15.270. Validation errors are equally high across its eight runs (five on PEMS04, three on PEMS05), and checks of target units, masks, early-stopping histories, and score aggregation found no evaluation inconsistency. We therefore report these rows as a degenerate fit of the frozen configuration (learning rate 0.03, at most 100 epochs, patience 5), not as evidence about the method. The cause has not been isolated, and these results do not establish that TGCN generally fails on evolving sensors.}

\revisiontext{On PEMS04, EAC has the largest between-seed spread of the 13 methods ($24.505\pm2.596$): its five all-sensor MAEs are 22.568, 22.942, 22.449, 28.021, and 26.547. All five runs completed the same budget and evaluation protocol. This variability warrants further investigation but does not establish a causal explanation based on graph growth or initial backbone capacity. PEMS04's first benchmark graph already has 1,266 nodes. Online-NN's largest yearly seed spreads occur in 2003 and 2004, when there are no new-to-previous-year sensors and its protocol skips fitting. It also skips fitting in 2024. This behavior illustrates why cohort definitions and skipped-year policies must accompany aggregate scores.}

\subsubsection{RQ2: New-Node Adaptation Evaluation}
\label{sec:result_evo_newsensor}

\revisiontext{The PEMS04 rows of Table~\ref{tab:tsas_new_sensors_part1} average over 20 years with valid new-to-previous-year targets. The year 2002 defines the baseline catalog, while 2003, 2004, and 2024 have empty new-node cohorts. PECPM has the lowest 12-lead MAE ($17.315\pm0.093$), followed by Online-NN ($17.797\pm0.161$), Online-AN ($18.018\pm0.069$), and STRAP ($18.156\pm0.090$). Restricting the calculation to first-ever sensors preserves this order: $17.355\pm0.101$, $17.824\pm0.156$, $18.064\pm0.072$, and $18.207\pm0.092$. These are supervised adaptation results. Online-NN moves ahead of Online-AN in this cohort while trailing it on all sensors, so the two evaluation populations reveal different behavior even though PECPM leads both MAE summaries.}

\revisiontext{On PEMS05, the new-to-previous-year cohort has valid targets in 12 years. Online-AN has the lowest three-seed mean MAE ($9.984\pm0.034$), followed closely by PECPM ($10.002\pm0.054$), Online-NN ($10.044\pm0.055$), and STRAP ($10.202\pm0.052$). The Online-AN/Online-NN ordering changes across seeds: Online-NN is slightly lower at seed 47, whereas Online-AN is lower at seeds 48 and 49. Rankings also depend on the metric. First-ever scores and per-year cohort denominators remain separate in the released results.}

\subsubsection{RQ3: Variation across Districts, Seeds, and Calendar Periods}
\label{subsec:evo_variation}

\revisiontext{On PEMS05, Online-AN has the lowest 12-lead all-sensor MAE ($11.515\pm0.037$), followed by recent-only ASTGCN ($11.766\pm0.090$) and STRAP ($11.777\pm0.059$), while Retrain reaches $11.920\pm0.049$. PECPM reaches $13.467\pm0.276$ on all sensors despite its second-lowest new-node mean. Thus the PEMS04 ranking does not transfer unchanged to PEMS05, and restricting the population can materially change a method's position. We do not treat the near-tie between ASTGCN and STRAP on PEMS05 as significant.}

\revisiontext{Table~\ref{tab:tsas_three_seed_diagnostics} compares Pretrain, Retrain, Online-NN, Online-AN, TrafficStream, EAC, and ST-TTC in all nine districts. Among them, Online-AN has the lowest all-sensor mean MAE in eight districts. On PEMS12, EAC is lower ($15.117\pm0.217$ against $15.249\pm0.102$) and leads in two of the three seeds. On sensors new to the preceding year, Online-NN has the lowest mean in six districts (PEMS04, PEMS07, PEMS08, PEMS10, PEMS11, and PEMS12) and Online-AN in the other three, although Online-NN's all-sensor MAE is higher than Online-AN's for every district and seed. Online-AN improves on per-year Retrain for every seed in all 18 district--cohort cells of Figure~\ref{fig:district_paired_diffs}, whereas ST-TTC does so in 15. Pretrain, which fits only the first task, has the highest mean MAE of the seven in every district and cohort. On PEMS04, PECPM remains below all seven methods in both populations, so these seven-method orderings do not replace the 13-method comparison.}

\revisiontext{As a temporal diagnostic, we recompute equal-year means using only the 2018--2025 tasks from these same sequential runs. Catalogs grow proportionally less in this period, although PEMS05 still more than doubles (243 to 572 sensors). In every district of Figure~\ref{fig:district_paired_diffs}, Online-AN's all-sensor MAE over these years is lower than Retrain's for every seed. Its mean is 4.24\% lower on PEMS03, 3.24\% on PEMS04, 4.24\% on PEMS05, 4.91\% on PEMS06, 4.50\% on PEMS07, 3.67\% on PEMS08, 3.56\% on PEMS10, 4.22\% on PEMS11, and 3.83\% on PEMS12, so the observed advantage is not confined to the early years. Online-AN also has the lowest 2018--2025 all-sensor mean of the seven methods in every district, including PEMS12. Across all 13 methods on PEMS04 and PEMS05 (Table~\ref{tab:tsas_recent_slice}), PECPM, TrafficStream, and STKEC generally hold or improve their rank in this period, while EAC stays eleventh in every column. On PEMS05, PECPM's all-sensor MAE ($10.278\pm0.076$) nearly equals Online-AN's ($10.275\pm0.017$), moving from eighth over the full period to second, and TrafficStream moves from ninth to fourth, whereas ASTGCN falls from second to eighth. PECPM also has the lowest PEMS05 new-node MAE in this period ($9.069\pm0.139$). On PEMS04, PECPM leads both populations, as over the full period. This slice is descriptive: it follows each method's original reset or state-carryover policy and does not isolate graph growth as a cause, certify graph stability, or substitute for the controlled fixed- versus evolving-catalog comparison of Section~\ref{subsec:fixed_vs_evolving}.}

\begin{figure}[t]
  \centering
  \includegraphics[width=\linewidth]{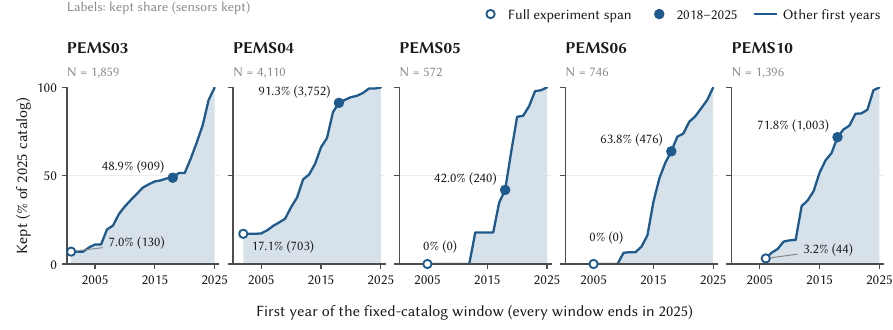}
  \Description{Five small line charts, one per district, of the share of the 2025 sensor catalog kept by a fixed intersection against the window's first year, with the full-span and 2018--2025 windows marked.}
  \caption{\revisiontext{Fixed-catalog coverage of each district's 2025 sensor set. A fixed-sensor catalog keeps only the sensors whose identities appear in every yearly catalog from the window's first year to 2025. Each panel plots the kept share of the district's 2025 catalog ($N$, under the district name) against that first year. The open marker is the full experiment span, which starts at the district's first experiment year (2001 for PEMS03, 2002 for PEMS04, 2005 for PEMS05 and PEMS06, and 2006 for PEMS10), and the filled marker is the 2018--2025 window. Labels give the kept share and, in parentheses, the number of sensors kept. The line joins the same calculation for every intermediate first year, from the same yearly catalogs, and passes through both windows. Under 2018--2025 the kept share ranges from 42.0\% (PEMS05, 240 of 572 sensors) to 91.3\% (PEMS04, 3,752 of 4,110). For PEMS05 and PEMS06 the full-span intersection is empty (0 of 572 and 0 of 746 sensors). This is a measured empty set, not a year of all-zero data. Coverage counts sensor identities only. It is neither a measurement-validity rate nor an accuracy gain.}}
  \label{fig:fixed_catalog_coverage}
\end{figure}

\revisiontext{A catalog-only check quantifies the coverage trade-off without asserting an accuracy gain (Figure~\ref{fig:fixed_catalog_coverage}, with counts in Table~\ref{tab:fixed_catalog_coverage}). Intersecting all yearly catalogs over the evaluated span leaves 130, 703, 0, 0, and 44 sensors for PEMS03, PEMS04, PEMS05, PEMS06, and PEMS10. Over 2018--2025, PEMS05 retains 240 common sensors compared with 572 in the 2025 catalog. A fixed intersection therefore excludes 332 of those 572 locations. This is a retrospective membership calculation: it does not establish observation completeness, strictly time-available sensor selection, or better predictions from using more sensors. The evolving setting exposes forecasts for the changing catalog and separately evaluates newly entering locations. Section~\ref{subsec:fixed_vs_evolving} compares fixed- and evolving-catalog training on these 240 PEMS05 sensors and on the 476 common PEMS06 sensors.}

\subsubsection{Sensitivity to Coordinate Availability}
\label{subsec:coordinate_sensitivity}

\revisiontext{We also rerun PEMS04 Online-AN over 2002--2025 with seeds 47--49, using only coordinate snapshots dated no later than each original yearly catalog anchor. Each sensor receives its earliest valid coordinate among the eligible frozen snapshots. Sensors without one have no spatial edges before the unchanged model-specific self-loop handling. This replaces the main construction's later-coordinate and centroid fallback policies. The Gaussian bandwidth is recomputed as the population standard deviation of positive pairwise distances among sensors with eligible coordinates, with the weight threshold kept at 0.1. Catalog identities, traffic values, observation masks, split boundaries, and training settings are unchanged. Each run starts a fresh 2002 model and maintains its own chronological checkpoint chain.}

\begin{figure}[t]
  \centering
  \includegraphics[width=\linewidth]{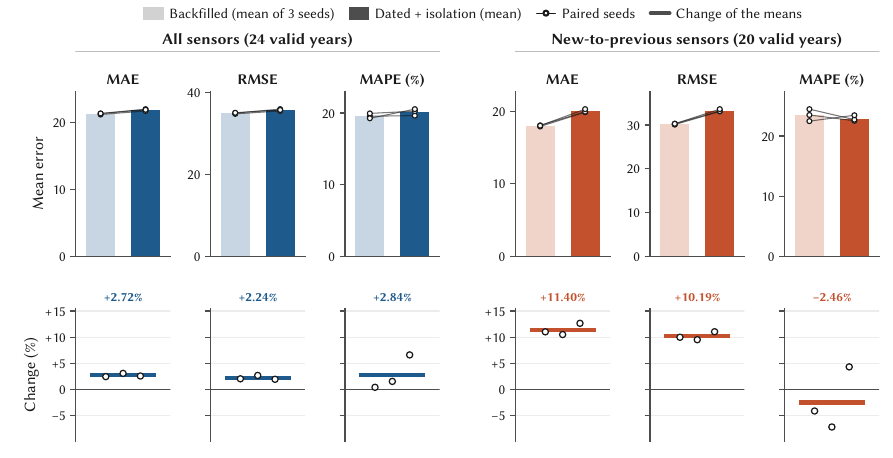}
  \Description{Six panels for MAE, RMSE and MAPE on all and new-to-previous sensors of PEMS04: bars compare the three-seed means under backfilled and dated coordinates, and a lower row shows each seed's relative change.}
  \caption{\revisiontext{Sensitivity of Online-AN on PEMS04 to the coordinate policy (matched seeds 47--49, full 12-lead horizon). Top row: the three-seed mean of each metric under the main comparison's retrospective coordinates (the baseline, shown as the light bar labeled Backfilled) and under dated coordinates with isolation of sensors that lack a dated position (dark bar). Bars start at zero, and the thin lines join each seed's two scores. Bottom row, on one axis shared by all panels: each seed's relative change (open circles) and the relative change of the means (thick line and label), $100\,(\bar{x}_{\mathrm{dated}}/\bar{x}_{\mathrm{backfilled}}-1)$, which is not the average of the per-seed changes. Left: all sensors (24 valid years). Right: new-to-previous sensors (20 valid years). Each top panel has its own scale, the two cohorts' absolute errors are not a measure of their difficulty, and because MAPE is itself a percentage its change is relative, not in percentage points. MAE and RMSE rise for every seed in both cohorts (all-sensor MAE +2.72\%, new-to-previous MAE +11.40\%), whereas new-to-previous MAPE falls for 2 seeds and rises for 1 (change of means $-$2.46\%). The coordinate dating and the handling of missing coordinates change together while the retrospective sensor catalog is unchanged, so this is a sensitivity check, not a prospective-deployment validation or a graph-robustness ranking of all methods. Exact values are in Table~\ref{tab:tsas_coordinate_sensitivity}.}}
  \label{fig:coordinate_sensitivity}
\end{figure}

\revisiontext{Figure~\ref{fig:coordinate_sensitivity} compares these runs with the same three seeds of the retrospective main comparison, and Table~\ref{tab:tsas_coordinate_sensitivity} lists the values. All-sensor MAE increases from $21.346\pm0.083$ to $21.926\pm0.139$ (+2.72\%), and new-to-previous-year MAE from $17.988\pm0.078$ to $20.038\pm0.225$ (+11.40\%). Each seed shows an MAE increase for both populations. RMSE also increases, whereas mean new-node MAPE decreases, so the policy does not affect every metric in the same direction. Evaluation populations match within each seed pair.}

\revisiontext{This is a joint sensitivity to metadata timing and missing-location handling at a fixed retrospective catalog. It does not isolate either change, establish strictly prospective catalog availability, or test whether the 13-method ranking persists. These runs used the four PEMS04 metadata snapshots of the main graph construction (2001-08-01, 2004-03-25, 2014-05-09, and 2024-03-20). With them, 1,898 of 2,750 catalog sensors lack eligible coordinates at the 2014 anchor. The archive also holds 211 non-empty dated PEMS04 metadata snapshots from 2001 to 2026, released with the coordinate files, and with all of them every 2014 catalog sensor has an eligible coordinate. The runs therefore isolate far more sensors than complete dated metadata would require, and their effect sizes should not be read as the effect of metadata timing alone. The larger new-node MAE change shows why graph-construction provenance should accompany benchmark results.}

\subsubsection{Fixed- versus Evolving-Catalog Training}
\label{subsec:fixed_vs_evolving}

\revisiontext{To test whether training on the evolving catalog improves accuracy on long-lived sensors, we add a matched control on PEMS05 and PEMS06 over 2018--2025, the recent window of Figure~\ref{fig:fixed_catalog_coverage}. The fixed set $S$ contains the sensors present in all eight yearly catalogs: 240 of the 572 PEMS05 sensors and 476 of the 746 PEMS06 sensors in 2025. Evolving-catalog training (E) uses each year's full catalog and archived graph, as in the main comparison. Fixed-catalog training (F) uses only $S$: its input normalization is recomputed from the accepted training values of $S$, its graph is the induced subgraph of the archived yearly graph with unchanged edge weights, and its early stopping uses validation targets on $S$. Retrain and Online-AN run with seeds 47--49 under the frozen protocol and configurations. Both start from new models in 2018, and Online-AN then fine-tunes year by year within its own chain. E and F are scored on the same accepted test targets of $S$, and the 24 runs cover 192 method--seed--year tasks.}

\revisiontext{Table~\ref{tab:tsas_fixed_vs_evolving} shows no consistent accuracy difference on $S$. On PEMS05, E has a lower mean MAE than F with both methods (paired E$-$F: $-0.096\pm0.041$ for Retrain and $-0.087\pm0.180$ for Online-AN) and a mean MAPE about 0.96 percentage points lower. On PEMS06, E has a slightly higher mean MAE ($+0.025\pm0.063$ and $+0.110\pm0.115$). Each mean MAE difference is below 1\% of the error level, and its sign holds for all three seeds only for PEMS05 Retrain (E lower) and PEMS06 Online-AN (F lower). Training on the evolving catalog therefore neither consistently improves nor degrades forecasts for the long-lived sensors, while the same models also forecast the sensors outside $S$: 58\% of the 2025 catalog on PEMS05 and 36\% on PEMS06. As a consistency check, E-trained Retrain, which starts a new model every year, agrees with the main-comparison 2018--2025 all-sensor slice: MAE $10.717\pm0.041$ versus $10.730\pm0.037$ on PEMS05 and $13.380\pm0.115$ versus $13.360\pm0.047$ on PEMS06. Online-AN restarted in 2018 has a higher all-sensor MAE than the main-comparison runs trained since 2005 ($10.496\pm0.122$ versus $10.275\pm0.017$ on PEMS05 and $13.140\pm0.025$ versus $12.704\pm0.010$ on PEMS06), so this control isolates the training population, not the value of a longer history. Two districts, two methods, and three seeds do not support a significance claim.}

\revisiontext{\noindent\textbf{Interpretation.} PECPM leads both PEMS04 populations on full-horizon MAE, whereas Online-AN leads both on PEMS05. Across the nine districts, Online-AN leads the seven-method all-sensor comparison in eight and Online-NN the new-node comparison in six. Online fine-tuning remains a useful reference, but the evidence does not support calling it the best all-sensor method in every district. Training on the evolving catalog widens forecast coverage without a consistent accuracy change on long-lived sensors. Cohort-dependent ordering, metric choice, seed variability, and the documented implementation adaptations are essential to interpreting this benchmark. Error magnitudes from different target populations do not by themselves establish which population is intrinsically harder.}

\subsection{Computational Budget and Recorded Runtime}

\revisiontext{Table~\ref{tab:evo_runtime} reports complete-run wall times on PEMS04 for the seven methods timed on one GPU model (NVIDIA H200). These measurements include training, validation, testing, loading, and checkpoint I/O. Method-specific batch sizes, early stopping, and skipped fitting prevent an equal-compute interpretation. The other six methods ran on other or mixed GPU types and, in some cases, resumed from checkpoints, so we infer no common-hardware runtime ranking for them. The shared computational constraint is the stated epoch cap and stopping rule, not equal wall time.}

\begin{table}[ht]
\centering
\color{blue}
\small
\caption{\revisiontext{Complete PEMS04 2002--2025 wall time for the seven methods timed on NVIDIA H200 GPUs, in minutes (five-seed mean$\pm$sample standard deviation). Times include data loading, training, validation, checkpoint I/O, and testing but exclude scheduler queue time and pre-run checks.}}
\label{tab:evo_runtime}
\begin{tabular}{lr}
\toprule
Method & Minutes \\
\midrule
Pretrain & $5.5\pm0.2$ \\
Retrain & $104.3\pm1.3$ \\
Online-NN & $36.3\pm1.3$ \\
Online-AN & $46.4\pm2.6$ \\
TrafficStream & $39.9\pm1.6$ \\
EAC & $52.4\pm5.5$ \\
ST-TTC & $105.2\pm2.3$ \\
\bottomrule
\end{tabular}
\end{table}

\section{Conclusion}
\label{sec:proandcons}

\revisiontext{We introduced EvoXXLTraffic as a sensor-evolving extension of the XXLTraffic dataset family. Yearly catalogs and graph snapshots enable evaluation over changing sensor identities, with explicit masks, chronological splits, and traceable checkpoint transitions. The benchmark compares 13 baseline adaptations on PEMS04 with five seeds and PEMS05 with three seeds, and seven of them on the other seven districts with three seeds. PECPM has the lowest full-horizon mean MAE on both all-sensor and new-node populations on PEMS04, while \mbox{Online-AN} leads both on PEMS05. Among the seven methods, \mbox{Online-AN} has the lowest all-sensor mean MAE in eight of the nine districts, whereas Online-NN has the lowest new-node mean in six, so the leading method changes with the evaluation population. A matched control on PEMS05 and PEMS06 finds that training on the evolving catalog changes mean MAE on the sensors common to 2018--2025 by less than 1\% in either direction. The evolving setting thus widens coverage without a consistent accuracy gain or loss on long-lived sensors. These findings support reporting district, cohort, metric, and seed variation rather than a universal method ranking.}

\revisiontext{The evidence remains specific to the documented adaptations and frozen configurations, including recent-only ASTGCN and training-only adaptations of PECPM and STKEC. DCRNN, ASTGCN, TGCN, PECPM, STRAP, and STKEC are evaluated on PEMS04 and PEMS05 only. TGCN's high errors under its frozen configuration are reported, but their cause has not been established. Each yearly task covers 31 stored days. Accepted published flow can include upstream imputation, and the graph/catalog construction is retrospective. In the PEMS04 Online-AN sensitivity analysis, the dated-coordinate policy increases all-sensor MAE by 2.72\% and new-node MAE by 11.40\%, documenting the importance of graph provenance without isolating metadata timing from missing-location handling. The fixed-catalog control covers two districts, two methods, and three seeds, with models started in 2018. The inherited fixed-sensor results were not re-validated and are not reproduced here (Appendix~\ref{app:legacy_results}). Broader evidence is needed for claims about alternative graph constructions, measurement-quality policies, or strictly prospective deployment.}


\begin{acks}
This work was supported by the ARC Centre of Excellence for Automated Decision-Making and Society (CE200100005). We acknowledge the resources and services provided by the National Computational Infrastructure (NCI), which is supported by the Australian Government. This research was also partially supported by the ARC Training Centre for Whole Life Design of Carbon Neutral Infrastructure (IC230100015).
\end{acks}

\bibliographystyle{ACM-Reference-Format}
\bibliography{sample-base}

@String{Computing = "Computing" }

@String{Springer = "Springer-Verlag" }

@inproceedings{lai2018modeling,
  title={Modeling long- and short-term temporal patterns with deep neural networks},
  author={Lai, Guokun and Chang, Wei-Cheng and Yang, Yiming and Liu, Hanxiao},
  booktitle={The 41st International ACM SIGIR Conference on Research \& Development in Information Retrieval},
  pages={95--104},
  year={2018}
}

@inproceedings{song2020spatial,
  title={Spatial-temporal synchronous graph convolutional networks: A new framework for spatial-temporal network data forecasting},
  author={Song, Chao and Lin, Youfang and Guo, Shengnan and Wan, Huaiyu},
  booktitle={Proceedings of the AAAI Conference on Artificial Intelligence},
  volume={34},
  pages={914--921},
  year={2020}
}

@article{gu2023mamba,
  title={Mamba: Linear-time sequence modeling with selective state spaces},
  author={Gu, Albert and Dao, Tri},
  journal={arXiv preprint arXiv:2312.00752},
  year={2023}
}

@article{zhao2019t,
  title={T-GCN: A temporal graph convolutional network for traffic prediction},
  author={Zhao, Ling and Song, Yujiao and Zhang, Chao and Liu, Yu and Wang, Pu and Lin, Tao and Deng, Min and Li, Haifeng},
  journal={IEEE Transactions on Intelligent Transportation Systems},
  volume={21},
  number={9},
  pages={3848--3858},
  year={2019},
  publisher={IEEE}
}

@inproceedings{guo2019attention,
  title={Attention based spatial-temporal graph convolutional networks for traffic flow forecasting},
  author={Guo, Shengnan and Lin, Youfang and Feng, Ning and Song, Chao and Wan, Huaiyu},
  booktitle={Proceedings of the AAAI Conference on Artificial Intelligence},
  volume={33},
  pages={922--929},
  year={2019}
}

@article{zhang2025strap,
  title={{STRAP}: Spatio-temporal pattern retrieval for out-of-distribution generalization},
  author={Zhang, Haoyu and Zhang, Wentao and Miao, Hao and Jiang, Xinke and Fang, Yuchen and Zhang, Yifan},
  journal={Advances in Neural Information Processing Systems},
  volume={38},
  pages={118006--118041},
  year={2025},
  doi={10.52202/085713-3934}
}

@inproceedings{ma2025beyond,
  title={Beyond fixed variables: Expanding-variate time series forecasting via flat scheme and spatio-temporal focal learning},
  author={Ma, Minbo and Tang, Kai and Li, Huan and Teng, Fei and Zhang, Dalin and Li, Tianrui},
  booktitle={Proceedings of the 31st ACM SIGKDD Conference on Knowledge Discovery and Data Mining V. 2},
  pages={2054--2065},
  year={2025}
}

@inproceedings{chen2025expand,
  title={Expand and compress: Exploring tuning principles for continual spatio-temporal graph forecasting},
  author={Chen, Wei and Liang, Yuxuan},
  booktitle={International Conference on Learning Representations},
  year={2025}
}

@inproceedings{yin2025xxltraffic,
  title={XXLTraffic: Expanding and Extremely Long Traffic Forecasting beyond Test Adaptation},
  author={Yin, Du and Xue, Hao and Prabowo, Arian and Ao, Shuang and Salim, Flora},
  booktitle={Proceedings of the 33rd ACM International Conference on Advances in Geographic Information Systems},
  pages={511--521},
  year={2025},
  doi={10.1145/3748636.3762762}
}

@article{chen2025learning,
  title={Learning with calibration: Exploring test-time computing of spatio-temporal forecasting},
  author={Chen, Wei and Liang, Yuxuan},
  journal={Advances in Neural Information Processing Systems},
  volume={38},
  pages={155895--155929},
  year={2025},
  doi={10.52202/085713-5210}
}

@inproceedings{liugeneral,
  title={A General Spatio-Temporal Backbone with Scalable Contextual Pattern Bank for Urban Continual Forecasting},
  author={Liu, Aoyu and Zhang, Yaying},
  booktitle={The Fourteenth International Conference on Learning Representations},
  year={2026},
  url={https://openreview.net/forum?id=LHSea6DI8U}
}

@article{wang2023knowledge,
  title={Knowledge expansion and consolidation for continual traffic prediction with expanding graphs},
  author={Wang, Binwu and Zhang, Yudong and Shi, Jiahao and Wang, Pengkun and Wang, Xu and Bai, Lei and Wang, Yang},
  journal={IEEE Transactions on Intelligent Transportation Systems},
  volume={24},
  number={7},
  pages={7190--7201},
  year={2023},
  publisher={IEEE}
}

@inproceedings{wang2023pattern,
  title={Pattern expansion and consolidation on evolving graphs for continual traffic prediction},
  author={Wang, Binwu and Zhang, Yudong and Wang, Xu and Wang, Pengkun and Zhou, Zhengyang and Bai, Lei and Wang, Yang},
  booktitle={Proceedings of the 29th ACM SIGKDD Conference on Knowledge Discovery and Data Mining},
  pages={2223--2232},
  year={2023}
}

@inproceedings{zeng2023transformers,
  title={Are transformers effective for time series forecasting?},
  author={Zeng, Ailing and Chen, Muxi and Zhang, Lei and Xu, Qiang},
  booktitle={Proceedings of the AAAI Conference on Artificial Intelligence},
  volume={37},
  pages={11121--11128},
  year={2023}
}

@inproceedings{wu2022timesnet,
  title={TimesNet: Temporal 2D-variation modeling for general time series analysis},
  author={Wu, Haixu and Hu, Tengge and Liu, Yong and Zhou, Hang and Wang, Jianmin and Long, Mingsheng},
  booktitle={The Eleventh International Conference on Learning Representations},
  year={2023}
}

@inproceedings{zhou2022fedformer,
  title={FEDformer: Frequency enhanced decomposed transformer for long-term series forecasting},
  author={Zhou, Tian and Ma, Ziqing and Wen, Qingsong and Wang, Xue and Sun, Liang and Jin, Rong},
  booktitle={International Conference on Machine Learning},
  pages={27268--27286},
  year={2022},
  organization={PMLR}
}

@article{chen2001freeway,
  title={Freeway performance measurement system: mining loop detector data},
  author={Chen, Chao and Petty, Karl and Skabardonis, Alexander and Varaiya, Pravin and Jia, Zhanfeng},
  journal={Transportation Research Record},
  volume={1748},
  pages={96--102},
  year={2001},
  publisher={SAGE Publications Sage CA: Los Angeles, CA}
}

@article{wu2021autoformer,
  title={Autoformer: Decomposition transformers with auto-correlation for long-term series forecasting},
  author={Wu, Haixu and Xu, Jiehui and Wang, Jianmin and Long, Mingsheng},
  journal={Advances in Neural Information Processing Systems},
  volume={34},
  pages={22419--22430},
  year={2021}
}

@inproceedings{wang2023micn,
  title={MICN: Multi-scale local and global context modeling for long-term series forecasting},
  author={Wang, Huiqiang and Peng, Jian and Huang, Feihu and Wang, Jince and Chen, Junhui and Xiao, Yifei},
  booktitle={The Eleventh International Conference on Learning Representations},
  year={2023}
}

@inproceedings{zhou2021informer,
  title={Informer: Beyond efficient transformer for long sequence time-series forecasting},
  author={Zhou, Haoyi and Zhang, Shanghang and Peng, Jieqi and Zhang, Shuai and Li, Jianxin and Xiong, Hui and Zhang, Wancai},
  booktitle={Proceedings of the AAAI Conference on Artificial Intelligence},
  volume={35},
  pages={11106--11115},
  year={2021}
}

@inproceedings{jiang2023pdformer,
  title={PDFormer: Propagation delay-aware dynamic long-range transformer for traffic flow prediction},
  author={Jiang, Jiawei and Han, Chengkai and Zhao, Wayne Xin and Wang, Jingyuan},
  booktitle={Proceedings of the AAAI Conference on Artificial Intelligence},
  volume={37},
  pages={4365--4373},
  year={2023}
}

@article{bai2020adaptive,
  title={Adaptive graph convolutional recurrent network for traffic forecasting},
  author={Bai, Lei and Yao, Lina and Li, Can and Wang, Xianzhi and Wang, Can},
  journal={Advances in Neural Information Processing Systems},
  volume={33},
  pages={17804--17815},
  year={2020}
}

@inproceedings{nietime,
  title={A Time Series is Worth 64 Words: Long-term Forecasting with Transformers},
  author={Nie, Yuqi and Nguyen, Nam H and Sinthong, Phanwadee and Kalagnanam, Jayant},
  booktitle={The Eleventh International Conference on Learning Representations},
  year={2023},
  url={https://openreview.net/forum?id=Jbdc0vTOcol}
}

@inproceedings{wu2019graph,
  title={Graph WaveNet for deep spatial-temporal graph modeling},
  author={Wu, Zonghan and Pan, Shirui and Long, Guodong and Jiang, Jing and Zhang, Chengqi},
  booktitle={Proceedings of the 28th International Joint Conference on Artificial Intelligence},
  pages={1907--1913},
  year={2019}
}

@inproceedings{yu2018spatio,
  title={Spatio-temporal graph convolutional networks: a deep learning framework for traffic forecasting},
  author={Yu, Bing and Yin, Haoteng and Zhu, Zhanxing},
  booktitle={Proceedings of the 27th International Joint Conference on Artificial Intelligence},
  pages={3634--3640},
  year={2018}
}

@inproceedings{li2018diffusion,
  title={Diffusion Convolutional Recurrent Neural Network: Data-Driven Traffic Forecasting},
  author={Li, Yaguang and Yu, Rose and Shahabi, Cyrus and Liu, Yan},
  booktitle={International Conference on Learning Representations},
  year={2018}
}

@inproceedings{liu2024itransformer,
title={iTransformer: Inverted Transformers Are Effective for Time Series Forecasting},
author={Yong Liu and Tengge Hu and Haoran Zhang and Haixu Wu and Shiyu Wang and Lintao Ma and Mingsheng Long},
booktitle={The Twelfth International Conference on Learning Representations},
year={2024},
url={https://openreview.net/forum?id=JePfAI8fah}
}

@article{prabowo2024traffic,
  title={Traffic forecasting on new roads using spatial contrastive pre-training (SCPT)},
  author={Prabowo, Arian and Xue, Hao and Shao, Wei and Koniusz, Piotr and Salim, Flora D},
  journal={Data Mining and Knowledge Discovery},
  volume={38},
  pages={913--937},
  year={2024},
  publisher={Springer}
}

@article{liu2024largest,
  title={{LargeST}: A Benchmark Dataset for Large-Scale Traffic Forecasting},
  author={Liu, Xu and Xia, Yutong and Liang, Yuxuan and Hu, Junfeng and Wang, Yiwei and Bai, Lei and Huang, Chao and Liu, Zhenguang and Hooi, Bryan and Zimmermann, Roger},
  journal={Advances in Neural Information Processing Systems},
  volume={36},
  year={2023}
}

@inproceedings{ijcai2021p0498,
  title     = {TrafficStream: A Streaming Traffic Flow Forecasting Framework Based on Graph Neural Networks and Continual Learning},
  author    = {Chen, Xu and Wang, Junshan and Xie, Kunqing},
  booktitle = {Proceedings of the Thirtieth International Joint Conference on
               Artificial Intelligence, {IJCAI-21}},
  publisher = {International Joint Conferences on Artificial Intelligence Organization},
  editor    = {Zhi-Hua Zhou},
  pages     = {3620--3626},
  year      = {2021},
  month     = {8},
  note      = {Main Track},
  doi       = {10.24963/ijcai.2021/498},
  url       = {https://doi.org/10.24963/ijcai.2021/498},
}

\appendix
\raggedbottom
\renewcommand{\arraystretch}{0.86}

\section{Detailed Sensor-Evolving Results}
\label{app:evo_details}
\revisiontext{These tables expand the main-text results without adding runs. Tables~\ref{tab:tsas_three_seed_diagnostics}, \ref{tab:fixed_catalog_coverage}, and~\ref{tab:tsas_coordinate_sensitivity} list the values underlying Figures~\ref{fig:district_paired_diffs}, \ref{fig:fixed_catalog_coverage}, and~\ref{fig:coordinate_sensitivity}, Table~\ref{tab:tsas_naive_references} gives the training-free references at every horizon and metric, Table~\ref{tab:tsas_recent_slice} the 2018--2025 slice of all 13 methods, and Table~\ref{tab:tsas_fixed_vs_evolving} the fixed- versus evolving-catalog control of Section~\ref{subsec:fixed_vs_evolving}. Method names denote the executed adaptations in Section~\ref{subsec:baseline_scope}.}
\section{Inherited Fixed-Sensor Results}
\label{app:legacy_results}
\revisiontext{The conference paper~\cite{yin2025xxltraffic} reports the inherited fixed-sensor experiments. They cover gap forecasting with 1-, 1.5-, and 2-year gaps on PEMS03, PEMS04, PEMS08, NSW, and six further PeMS districts, together with hourly and daily forecasting, a MICN data-fraction comparison, a DLinear input-length comparison, and per-epoch timings. They evaluated the long-term forecasting models Informer~\cite{zhou2021informer}, MICN~\cite{wang2023micn}, FEDformer~\cite{zhou2022fedformer}, PatchTST~\cite{nietime}, Autoformer~\cite{wu2021autoformer}, iTransformer~\cite{liu2024itransformer}, DLinear~\cite{zeng2023transformers}, and Mamba~\cite{gu2023mamba}, implemented with the Time-Series-Library~\cite{wu2022timesnet}. They also evaluated the traffic models STGCN~\cite{yu2018spatio}, ASTGCN~\cite{guo2019attention}, GWN~\cite{wu2019graph}, AGCRN~\cite{bai2020adaptive}, and PDFormer~\cite{jiang2023pdformer}. The settings remain part of the XXLTraffic dataset family (Section~\ref{subsec:fixed_protocols}). Their results were not re-validated under the corrected protocol of Section~\ref{subsec:evo_streaming_protocol}, so this article refers to the conference paper for them instead of reproducing them. The sensor-evolving benchmark in Sections~\ref{sec:benchmark_protocols} and~\ref{exp_section6} does not depend on them.}

\begin{table}[p]
\centering
\color{blue}
\caption{\revisiontext{All-sensor results for the two complete 13-method comparisons. PEMS04 (2002--2025) uses seeds 47--51, and PEMS05 (2005--2025) uses seeds 47--49. Values are mean$\pm$sample standard deviation after equal-year aggregation. Horizons 3/6/12 pool leads 1--3/1--6/1--12, and Avg equals the 12-lead pool. Bold and underline indicate the two lowest unrounded means, without a significance claim. ASTGCN is recent-only, and TGCN uses the frozen configuration discussed in Section~\ref{sec:result_evo_allsensor}. Table~\ref{tab:tsas_three_seed_diagnostics} reports seven of these methods in all nine districts.}}
\label{tab:tsas_evo_part1}
\resizebox{\textwidth}{!}{%
\begin{tabular}{ccc|ccc|cccc|cccc|cc}
\toprule
\multirow{2}{*}{\textbf{Dataset}} & \multirow{2}{*}{\textbf{Metric}} & \multirow{2}{*}{\textbf{Horizon}}
& \multicolumn{3}{c|}{\textbf{Static adaptations}}
& \multicolumn{4}{c|}{\textbf{Training schemes}}
& \multicolumn{4}{c|}{\textbf{Continual adaptations}}
& \multicolumn{2}{c}{\textbf{Retrieval/calibration}} \\
\cmidrule(lr){4-6} \cmidrule(lr){7-10} \cmidrule(lr){11-14} \cmidrule(lr){15-16}
 & & & \textbf{DCRNN} & \textbf{ASTGCN} & \textbf{TGCN}
 & \textbf{Pretrain} & \textbf{Retrain} & \textbf{Online-NN} & \textbf{Online-AN}
 & \textbf{TrafficStream} & \textbf{PECPM} & \textbf{STKEC} & \textbf{EAC}
 & \textbf{STRAP} & \textbf{ST-TTC} \\
\midrule
\multirow{12}{*}{\textbf{PEMS04}} & \multirow{4}{*}{MAE} & 3 & 17.71$_{\pm0.30}$ & 17.74$_{\pm0.05}$ & 130.52$_{\pm2.16}$ & 38.87$_{\pm1.70}$ & 17.06$_{\pm0.06}$ & 19.10$_{\pm0.85}$ & \cellcolor{gray!18}\textbf{16.68$_{\pm0.04}$} & 17.18$_{\pm0.20}$ & \underline{16.85$_{\pm0.13}$} & 17.80$_{\pm0.29}$ & 20.31$_{\pm1.70}$ & 16.93$_{\pm0.09}$ & 17.02$_{\pm0.06}$ \\
 &  & 6 & 20.00$_{\pm0.29}$ & 19.48$_{\pm0.06}$ & 130.51$_{\pm2.16}$ & 39.68$_{\pm1.79}$ & 18.73$_{\pm0.05}$ & 20.66$_{\pm0.88}$ & \underline{18.29$_{\pm0.05}$} & 18.78$_{\pm0.21}$ & \cellcolor{gray!18}\textbf{18.21$_{\pm0.09}$} & 19.28$_{\pm0.17}$ & 21.42$_{\pm1.92}$ & 18.49$_{\pm0.08}$ & 18.67$_{\pm0.05}$ \\
 &  & 12 & 24.46$_{\pm0.39}$ & 23.01$_{\pm0.12}$ & 130.50$_{\pm2.16}$ & 42.10$_{\pm1.61}$ & 22.08$_{\pm0.09}$ & 23.74$_{\pm0.83}$ & \underline{21.39$_{\pm0.09}$} & 22.01$_{\pm0.37}$ & \cellcolor{gray!18}\textbf{20.76$_{\pm0.08}$} & 22.52$_{\pm0.19}$ & 24.51$_{\pm2.60}$ & 21.57$_{\pm0.09}$ & 21.96$_{\pm0.09}$ \\
 &  & Avg & 24.46$_{\pm0.39}$ & 23.01$_{\pm0.12}$ & 130.50$_{\pm2.16}$ & 42.10$_{\pm1.61}$ & 22.08$_{\pm0.09}$ & 23.74$_{\pm0.83}$ & \underline{21.39$_{\pm0.09}$} & 22.01$_{\pm0.37}$ & \cellcolor{gray!18}\textbf{20.76$_{\pm0.08}$} & 22.52$_{\pm0.19}$ & 24.51$_{\pm2.60}$ & 21.57$_{\pm0.09}$ & 21.96$_{\pm0.09}$ \\
\cmidrule(lr){2-16}
 & \multirow{4}{*}{RMSE} & 3 & 28.41$_{\pm0.21}$ & 29.02$_{\pm0.09}$ & 164.62$_{\pm2.36}$ & 48.27$_{\pm2.02}$ & 27.89$_{\pm0.07}$ & 29.97$_{\pm0.81}$ & \cellcolor{gray!18}\textbf{27.45$_{\pm0.07}$} & 28.02$_{\pm0.22}$ & \underline{27.63$_{\pm0.15}$} & 28.63$_{\pm0.26}$ & 30.87$_{\pm1.86}$ & 27.68$_{\pm0.10}$ & 27.85$_{\pm0.06}$ \\
 &  & 6 & 31.91$_{\pm0.25}$ & 31.88$_{\pm0.11}$ & 164.61$_{\pm2.35}$ & 49.70$_{\pm2.14}$ & 30.75$_{\pm0.06}$ & 32.67$_{\pm0.85}$ & \underline{30.16$_{\pm0.10}$} & 30.77$_{\pm0.28}$ & \cellcolor{gray!18}\textbf{30.03$_{\pm0.12}$} & 31.27$_{\pm0.16}$ & 32.97$_{\pm2.11}$ & 30.35$_{\pm0.09}$ & 30.67$_{\pm0.07}$ \\
 &  & 12 & 38.64$_{\pm0.54}$ & 37.31$_{\pm0.20}$ & 164.60$_{\pm2.34}$ & 53.68$_{\pm1.93}$ & 36.14$_{\pm0.13}$ & 37.72$_{\pm0.77}$ & \underline{35.07$_{\pm0.16}$} & 36.02$_{\pm0.60}$ & \cellcolor{gray!18}\textbf{34.23$_{\pm0.15}$} & 36.65$_{\pm0.30}$ & 37.94$_{\pm3.10}$ & 35.29$_{\pm0.13}$ & 35.99$_{\pm0.14}$ \\
 &  & Avg & 38.64$_{\pm0.54}$ & 37.31$_{\pm0.20}$ & 164.60$_{\pm2.34}$ & 53.68$_{\pm1.93}$ & 36.14$_{\pm0.13}$ & 37.72$_{\pm0.77}$ & \underline{35.07$_{\pm0.16}$} & 36.02$_{\pm0.60}$ & \cellcolor{gray!18}\textbf{34.23$_{\pm0.15}$} & 36.65$_{\pm0.30}$ & 37.94$_{\pm3.10}$ & 35.29$_{\pm0.13}$ & 35.99$_{\pm0.14}$ \\
\cmidrule(lr){2-16}
 & \multirow{4}{*}{MAPE (\%)} & 3 & 19.64$_{\pm1.27}$ & 18.89$_{\pm0.30}$ & 353.14$_{\pm5.04}$ & 134.80$_{\pm12.18}$ & 16.48$_{\pm0.12}$ & 17.84$_{\pm1.08}$ & \cellcolor{gray!18}\textbf{15.36$_{\pm0.17}$} & 16.44$_{\pm0.95}$ & \underline{15.82$_{\pm0.26}$} & 17.68$_{\pm0.48}$ & 32.37$_{\pm22.38}$ & 16.54$_{\pm0.68}$ & 16.30$_{\pm0.11}$ \\
 &  & 6 & 22.02$_{\pm1.09}$ & 20.69$_{\pm0.42}$ & 353.37$_{\pm5.10}$ & 137.37$_{\pm12.95}$ & 17.90$_{\pm0.16}$ & 19.08$_{\pm1.03}$ & \underline{16.69$_{\pm0.21}$} & 17.74$_{\pm0.87}$ & \cellcolor{gray!18}\textbf{16.66$_{\pm0.19}$} & 18.64$_{\pm0.47}$ & 33.58$_{\pm23.60}$ & 17.64$_{\pm0.68}$ & 17.69$_{\pm0.16}$ \\
 &  & 12 & 26.38$_{\pm0.99}$ & 25.29$_{\pm0.53}$ & 353.97$_{\pm5.08}$ & 142.56$_{\pm13.05}$ & 21.22$_{\pm0.31}$ & 22.01$_{\pm1.04}$ & \underline{19.66$_{\pm0.26}$} & 20.92$_{\pm0.96}$ & \cellcolor{gray!18}\textbf{18.61$_{\pm0.14}$} & 21.20$_{\pm0.51}$ & 37.13$_{\pm26.10}$ & 20.29$_{\pm0.72}$ & 20.95$_{\pm0.30}$ \\
 &  & Avg & 26.38$_{\pm0.99}$ & 25.29$_{\pm0.53}$ & 353.97$_{\pm5.08}$ & 142.56$_{\pm13.05}$ & 21.22$_{\pm0.31}$ & 22.01$_{\pm1.04}$ & \underline{19.66$_{\pm0.26}$} & 20.92$_{\pm0.96}$ & \cellcolor{gray!18}\textbf{18.61$_{\pm0.14}$} & 21.20$_{\pm0.51}$ & 37.13$_{\pm26.10}$ & 20.29$_{\pm0.72}$ & 20.95$_{\pm0.30}$ \\
\midrule
\multirow{12}{*}{\textbf{PEMS05}} & \multirow{4}{*}{MAE} & 3 & 10.14$_{\pm0.13}$ & 10.09$_{\pm0.05}$ & 50.37$_{\pm0.31}$ & 25.97$_{\pm0.18}$ & 9.90$_{\pm0.02}$ & 11.71$_{\pm0.43}$ & \cellcolor{gray!18}\textbf{9.71$_{\pm0.01}$} & 11.54$_{\pm0.15}$ & 11.60$_{\pm0.28}$ & 10.11$_{\pm0.08}$ & 12.56$_{\pm0.53}$ & 9.93$_{\pm0.03}$ & \underline{9.89$_{\pm0.02}$} \\
 &  & 6 & 10.93$_{\pm0.18}$ & 10.64$_{\pm0.06}$ & 50.41$_{\pm0.31}$ & 26.34$_{\pm0.20}$ & 10.57$_{\pm0.03}$ & 12.42$_{\pm0.39}$ & \cellcolor{gray!18}\textbf{10.33$_{\pm0.01}$} & 12.25$_{\pm0.08}$ & 12.22$_{\pm0.27}$ & 10.74$_{\pm0.06}$ & 12.91$_{\pm0.51}$ & \underline{10.53$_{\pm0.04}$} & 10.56$_{\pm0.02}$ \\
 &  & 12 & 12.48$_{\pm0.29}$ & \underline{11.77$_{\pm0.09}$} & 50.52$_{\pm0.32}$ & 27.44$_{\pm0.17}$ & 11.92$_{\pm0.05}$ & 13.96$_{\pm0.32}$ & \cellcolor{gray!18}\textbf{11.52$_{\pm0.04}$} & 13.71$_{\pm0.05}$ & 13.47$_{\pm0.28}$ & 12.11$_{\pm0.06}$ & 14.06$_{\pm0.59}$ & 11.78$_{\pm0.06}$ & 11.90$_{\pm0.04}$ \\
 &  & Avg & 12.48$_{\pm0.29}$ & \underline{11.77$_{\pm0.09}$} & 50.52$_{\pm0.32}$ & 27.44$_{\pm0.17}$ & 11.92$_{\pm0.05}$ & 13.96$_{\pm0.32}$ & \cellcolor{gray!18}\textbf{11.52$_{\pm0.04}$} & 13.71$_{\pm0.05}$ & 13.47$_{\pm0.28}$ & 12.11$_{\pm0.06}$ & 14.06$_{\pm0.59}$ & 11.78$_{\pm0.06}$ & 11.90$_{\pm0.04}$ \\
\cmidrule(lr){2-16}
 & \multirow{4}{*}{RMSE} & 3 & 15.97$_{\pm0.15}$ & 16.21$_{\pm0.10}$ & 67.84$_{\pm0.08}$ & 36.81$_{\pm0.27}$ & 15.95$_{\pm0.02}$ & 18.95$_{\pm0.77}$ & \cellcolor{gray!18}\textbf{15.64$_{\pm0.03}$} & 18.77$_{\pm0.31}$ & 18.96$_{\pm0.49}$ & 16.18$_{\pm0.07}$ & 19.00$_{\pm0.68}$ & \underline{15.90$_{\pm0.02}$} & 15.93$_{\pm0.02}$ \\
 &  & 6 & 17.28$_{\pm0.20}$ & 17.19$_{\pm0.11}$ & 67.88$_{\pm0.09}$ & 37.54$_{\pm0.26}$ & 17.21$_{\pm0.03}$ & 20.32$_{\pm0.69}$ & \cellcolor{gray!18}\textbf{16.75$_{\pm0.03}$} & 20.11$_{\pm0.21}$ & 20.20$_{\pm0.48}$ & 17.40$_{\pm0.05}$ & 19.60$_{\pm0.54}$ & \underline{17.04$_{\pm0.04}$} & 17.19$_{\pm0.04}$ \\
 &  & 12 & 19.80$_{\pm0.40}$ & \underline{19.00$_{\pm0.14}$} & 68.05$_{\pm0.11}$ & 39.32$_{\pm0.17}$ & 19.58$_{\pm0.08}$ & 23.06$_{\pm0.55}$ & \cellcolor{gray!18}\textbf{18.78$_{\pm0.07}$} & 22.68$_{\pm0.12}$ & 22.46$_{\pm0.56}$ & 19.83$_{\pm0.09}$ & 21.48$_{\pm0.83}$ & 19.23$_{\pm0.10}$ & 19.56$_{\pm0.07}$ \\
 &  & Avg & 19.80$_{\pm0.40}$ & \underline{19.00$_{\pm0.14}$} & 68.05$_{\pm0.11}$ & 39.32$_{\pm0.17}$ & 19.58$_{\pm0.08}$ & 23.06$_{\pm0.55}$ & \cellcolor{gray!18}\textbf{18.78$_{\pm0.07}$} & 22.68$_{\pm0.12}$ & 22.46$_{\pm0.56}$ & 19.83$_{\pm0.09}$ & 21.48$_{\pm0.83}$ & 19.23$_{\pm0.10}$ & 19.56$_{\pm0.07}$ \\
\cmidrule(lr){2-16}
 & \multirow{4}{*}{MAPE (\%)} & 3 & 26.09$_{\pm2.00}$ & 23.42$_{\pm0.56}$ & 274.07$_{\pm8.35}$ & 118.33$_{\pm1.42}$ & 21.61$_{\pm0.19}$ & 25.11$_{\pm1.32}$ & \cellcolor{gray!18}\textbf{21.19$_{\pm0.11}$} & 24.50$_{\pm0.86}$ & 25.44$_{\pm0.84}$ & 22.80$_{\pm0.80}$ & 42.41$_{\pm5.70}$ & 22.44$_{\pm0.19}$ & \underline{21.42$_{\pm0.18}$} \\
 &  & 6 & 27.99$_{\pm1.89}$ & 24.44$_{\pm0.54}$ & 274.52$_{\pm8.43}$ & 119.23$_{\pm1.37}$ & 22.89$_{\pm0.18}$ & 26.33$_{\pm1.33}$ & \cellcolor{gray!18}\textbf{22.28$_{\pm0.12}$} & 25.74$_{\pm1.15}$ & 26.22$_{\pm0.70}$ & 23.88$_{\pm0.58}$ & 43.52$_{\pm5.74}$ & 23.38$_{\pm0.22}$ & \underline{22.72$_{\pm0.18}$} \\
 &  & 12 & 31.56$_{\pm1.94}$ & 27.50$_{\pm0.63}$ & 275.50$_{\pm8.45}$ & 123.71$_{\pm1.98}$ & 25.79$_{\pm0.22}$ & 29.81$_{\pm1.02}$ & \cellcolor{gray!18}\textbf{24.69$_{\pm0.14}$} & 29.10$_{\pm1.58}$ & 28.49$_{\pm0.49}$ & 26.61$_{\pm0.30}$ & 45.89$_{\pm5.44}$ & 25.85$_{\pm0.39}$ & \underline{25.60$_{\pm0.31}$} \\
 &  & Avg & 31.56$_{\pm1.94}$ & 27.50$_{\pm0.63}$ & 275.50$_{\pm8.45}$ & 123.71$_{\pm1.98}$ & 25.79$_{\pm0.22}$ & 29.81$_{\pm1.02}$ & \cellcolor{gray!18}\textbf{24.69$_{\pm0.14}$} & 29.10$_{\pm1.58}$ & 28.49$_{\pm0.49}$ & 26.61$_{\pm0.30}$ & 45.89$_{\pm5.44}$ & 25.85$_{\pm0.39}$ & \underline{25.60$_{\pm0.31}$} \\
\bottomrule
\end{tabular}%
}
\end{table}

\begin{table}[p]
\centering
\color{blue}
\caption{\revisiontext{New-node adaptation errors for the two complete 13-method comparisons. The cohort is $\mathcal{V}_y\setminus\mathcal{V}_{y-1}$, including returning sensors. Every method except Pretrain has current-year training. PEMS04 has 20 eligible years and five seeds (47--51), and PEMS05 has 12 eligible years and three seeds (47--49). Years are weighted equally within a seed, and empty cohorts are omitted. Values are mean$\pm$sample standard deviation. Horizons are cumulative pools and Avg equals the 12-lead pool. Bold and underline indicate the two lowest unrounded means, without a significance claim. ASTGCN is recent-only, and TGCN uses the frozen configuration discussed in Section~\ref{sec:result_evo_allsensor}.}}
\label{tab:tsas_new_sensors_part1}
\resizebox{\textwidth}{!}{%
\begin{tabular}{ccc|ccc|cccc|cccc|cc}
\toprule
\multirow{2}{*}{\textbf{Dataset}} & \multirow{2}{*}{\textbf{Metric}} & \multirow{2}{*}{\textbf{Horizon}} & \multicolumn{3}{c|}{\textbf{Static adaptations}} & \multicolumn{4}{c|}{\textbf{Training schemes}} & \multicolumn{4}{c|}{\textbf{Continual adaptations}} & \multicolumn{2}{c}{\textbf{Retrieval/calibration}} \\
\cmidrule(lr){4-6} \cmidrule(lr){7-10} \cmidrule(lr){11-14} \cmidrule(lr){15-16}
 &  &  & \textbf{DCRNN} & \textbf{ASTGCN} & \textbf{TGCN} & \textbf{Pretrain} & \textbf{Retrain} & \textbf{Online-NN} & \textbf{Online-AN} & \textbf{TrafficStream} & \textbf{PECPM} & \textbf{STKEC} & \textbf{EAC} & \textbf{STRAP} & \textbf{ST-TTC} \\
\midrule
\multirow{12}{*}{\textbf{PEMS04}} & \multirow{4}{*}{MAE} & 3 & 15.03$_{\pm0.36}$ & 15.10$_{\pm0.06}$ & 129.70$_{\pm2.62}$ & 38.51$_{\pm1.67}$ & 14.28$_{\pm0.05}$ & \cellcolor{gray!18}\textbf{13.78$_{\pm0.07}$} & \underline{13.85$_{\pm0.05}$} & 14.08$_{\pm0.16}$ & 13.94$_{\pm0.11}$ & 14.73$_{\pm0.27}$ & 18.73$_{\pm2.49}$ & 14.08$_{\pm0.07}$ & 14.25$_{\pm0.04}$ \\
 &  & 6 & 17.14$_{\pm0.36}$ & 16.66$_{\pm0.07}$ & 129.70$_{\pm2.63}$ & 39.26$_{\pm1.79}$ & 15.73$_{\pm0.04}$ & \underline{15.14$_{\pm0.08}$} & 15.27$_{\pm0.06}$ & 15.47$_{\pm0.20}$ & \cellcolor{gray!18}\textbf{15.10$_{\pm0.09}$} & 16.01$_{\pm0.20}$ & 19.63$_{\pm2.71}$ & 15.43$_{\pm0.08}$ & 15.68$_{\pm0.03}$ \\
 &  & 12 & 21.26$_{\pm0.49}$ & 19.91$_{\pm0.13}$ & 129.70$_{\pm2.60}$ & 41.51$_{\pm1.72}$ & 18.69$_{\pm0.07}$ & \underline{17.80$_{\pm0.16}$} & 18.02$_{\pm0.07}$ & 18.29$_{\pm0.35}$ & \cellcolor{gray!18}\textbf{17.31$_{\pm0.09}$} & 18.81$_{\pm0.24}$ & 22.31$_{\pm3.31}$ & 18.16$_{\pm0.09}$ & 18.58$_{\pm0.07}$ \\
 &  & Avg & 21.26$_{\pm0.49}$ & 19.91$_{\pm0.13}$ & 129.70$_{\pm2.60}$ & 41.51$_{\pm1.72}$ & 18.69$_{\pm0.07}$ & \underline{17.80$_{\pm0.16}$} & 18.02$_{\pm0.07}$ & 18.29$_{\pm0.35}$ & \cellcolor{gray!18}\textbf{17.31$_{\pm0.09}$} & 18.81$_{\pm0.24}$ & 22.31$_{\pm3.31}$ & 18.16$_{\pm0.09}$ & 18.58$_{\pm0.07}$ \\
\cmidrule(lr){2-16}
 & \multirow{4}{*}{RMSE} & 3 & 24.37$_{\pm0.24}$ & 24.88$_{\pm0.09}$ & 159.37$_{\pm2.69}$ & 47.05$_{\pm1.95}$ & 23.91$_{\pm0.04}$ & \cellcolor{gray!18}\textbf{23.39$_{\pm0.08}$} & \underline{23.44$_{\pm0.06}$} & 23.69$_{\pm0.15}$ & 23.53$_{\pm0.11}$ & 24.34$_{\pm0.25}$ & 28.48$_{\pm2.73}$ & 23.62$_{\pm0.08}$ & 23.86$_{\pm0.03}$ \\
 &  & 6 & 27.58$_{\pm0.29}$ & 27.48$_{\pm0.11}$ & 159.37$_{\pm2.69}$ & 48.29$_{\pm2.00}$ & 26.42$_{\pm0.04}$ & \underline{25.74$_{\pm0.12}$} & 25.85$_{\pm0.08}$ & 26.12$_{\pm0.23}$ & \cellcolor{gray!18}\textbf{25.63$_{\pm0.10}$} & 26.67$_{\pm0.19}$ & 30.23$_{\pm2.94}$ & 25.99$_{\pm0.09}$ & 26.33$_{\pm0.03}$ \\
 &  & 12 & 33.88$_{\pm0.55}$ & 32.54$_{\pm0.23}$ & 159.37$_{\pm2.66}$ & 51.85$_{\pm1.75}$ & 31.24$_{\pm0.08}$ & \underline{30.02$_{\pm0.27}$} & 30.28$_{\pm0.10}$ & 30.74$_{\pm0.52}$ & \cellcolor{gray!18}\textbf{29.35$_{\pm0.12}$} & 31.38$_{\pm0.33}$ & 34.63$_{\pm3.70}$ & 30.42$_{\pm0.12}$ & 31.07$_{\pm0.08}$ \\
 &  & Avg & 33.88$_{\pm0.55}$ & 32.54$_{\pm0.23}$ & 159.37$_{\pm2.66}$ & 51.85$_{\pm1.75}$ & 31.24$_{\pm0.08}$ & \underline{30.02$_{\pm0.27}$} & 30.28$_{\pm0.10}$ & 30.74$_{\pm0.52}$ & \cellcolor{gray!18}\textbf{29.35$_{\pm0.12}$} & 31.38$_{\pm0.33}$ & 34.63$_{\pm3.70}$ & 30.42$_{\pm0.12}$ & 31.07$_{\pm0.08}$ \\
\cmidrule(lr){2-16}
 & \multirow{4}{*}{MAPE (\%)} & 3 & 24.14$_{\pm2.09}$ & 23.70$_{\pm1.13}$ & 530.07$_{\pm10.60}$ & 206.55$_{\pm18.38}$ & 20.20$_{\pm0.28}$ & \cellcolor{gray!18}\textbf{17.35$_{\pm0.33}$} & \underline{18.10$_{\pm0.51}$} & 18.97$_{\pm1.35}$ & 18.47$_{\pm0.36}$ & 21.11$_{\pm0.55}$ & 45.21$_{\pm39.95}$ & 19.82$_{\pm1.34}$ & 19.88$_{\pm0.24}$ \\
 &  & 6 & 27.34$_{\pm1.66}$ & 25.93$_{\pm1.34}$ & 530.48$_{\pm10.67}$ & 210.45$_{\pm19.73}$ & 21.89$_{\pm0.22}$ & \cellcolor{gray!18}\textbf{18.75$_{\pm0.35}$} & 19.72$_{\pm0.63}$ & 20.48$_{\pm1.29}$ & \underline{19.25$_{\pm0.27}$} & 22.16$_{\pm0.52}$ & 46.76$_{\pm41.83}$ & 21.10$_{\pm1.30}$ & 21.55$_{\pm0.17}$ \\
 &  & 12 & 32.93$_{\pm1.55}$ & 31.93$_{\pm1.28}$ & 531.39$_{\pm10.59}$ & 218.04$_{\pm20.88}$ & 26.12$_{\pm0.39}$ & \underline{22.20$_{\pm0.50}$} & 23.54$_{\pm0.72}$ & 24.33$_{\pm1.57}$ & \cellcolor{gray!18}\textbf{21.43$_{\pm0.28}$} & 25.04$_{\pm0.48}$ & 51.14$_{\pm45.53}$ & 24.33$_{\pm1.24}$ & 25.67$_{\pm0.36}$ \\
 &  & Avg & 32.93$_{\pm1.55}$ & 31.93$_{\pm1.28}$ & 531.39$_{\pm10.59}$ & 218.04$_{\pm20.88}$ & 26.12$_{\pm0.39}$ & \underline{22.20$_{\pm0.50}$} & 23.54$_{\pm0.72}$ & 24.33$_{\pm1.57}$ & \cellcolor{gray!18}\textbf{21.43$_{\pm0.28}$} & 25.04$_{\pm0.48}$ & 51.14$_{\pm45.53}$ & 24.33$_{\pm1.24}$ & 25.67$_{\pm0.36}$ \\
\midrule
\multirow{12}{*}{\textbf{PEMS05}} & \multirow{4}{*}{MAE} & 3 & 8.96$_{\pm0.35}$ & 8.78$_{\pm0.07}$ & 49.65$_{\pm4.20}$ & 23.66$_{\pm0.25}$ & 8.62$_{\pm0.06}$ & \underline{8.51$_{\pm0.02}$} & \cellcolor{gray!18}\textbf{8.42$_{\pm0.01}$} & 8.60$_{\pm0.04}$ & 8.72$_{\pm0.09}$ & 8.77$_{\pm0.11}$ & 11.10$_{\pm0.36}$ & 8.66$_{\pm0.05}$ & 8.61$_{\pm0.04}$ \\
 &  & 6 & 9.71$_{\pm0.40}$ & 9.31$_{\pm0.09}$ & 49.69$_{\pm4.21}$ & 23.97$_{\pm0.20}$ & 9.20$_{\pm0.05}$ & \underline{9.02$_{\pm0.01}$} & \cellcolor{gray!18}\textbf{8.95$_{\pm0.01}$} & 9.13$_{\pm0.04}$ & 9.17$_{\pm0.06}$ & 9.29$_{\pm0.07}$ & 11.43$_{\pm0.32}$ & 9.17$_{\pm0.05}$ & 9.18$_{\pm0.04}$ \\
 &  & 12 & 11.12$_{\pm0.51}$ & 10.38$_{\pm0.15}$ & 49.79$_{\pm4.17}$ & 24.95$_{\pm0.26}$ & 10.37$_{\pm0.05}$ & 10.04$_{\pm0.06}$ & \cellcolor{gray!18}\textbf{9.98$_{\pm0.03}$} & 10.23$_{\pm0.05}$ & \underline{10.00$_{\pm0.05}$} & 10.41$_{\pm0.04}$ & 12.44$_{\pm0.40}$ & 10.20$_{\pm0.05}$ & 10.34$_{\pm0.05}$ \\
 &  & Avg & 11.12$_{\pm0.51}$ & 10.38$_{\pm0.15}$ & 49.79$_{\pm4.17}$ & 24.95$_{\pm0.26}$ & 10.37$_{\pm0.05}$ & 10.04$_{\pm0.06}$ & \cellcolor{gray!18}\textbf{9.98$_{\pm0.03}$} & 10.23$_{\pm0.05}$ & \underline{10.00$_{\pm0.05}$} & 10.41$_{\pm0.04}$ & 12.44$_{\pm0.40}$ & 10.20$_{\pm0.05}$ & 10.34$_{\pm0.05}$ \\
\cmidrule(lr){2-16}
 & \multirow{4}{*}{RMSE} & 3 & 13.83$_{\pm0.37}$ & 13.84$_{\pm0.09}$ & 62.25$_{\pm3.77}$ & 31.08$_{\pm0.21}$ & 13.74$_{\pm0.03}$ & \underline{13.53$_{\pm0.05}$} & \cellcolor{gray!18}\textbf{13.45$_{\pm0.01}$} & 13.63$_{\pm0.08}$ & 13.85$_{\pm0.09}$ & 13.93$_{\pm0.12}$ & 16.37$_{\pm0.57}$ & 13.74$_{\pm0.07}$ & 13.73$_{\pm0.04}$ \\
 &  & 6 & 15.03$_{\pm0.43}$ & 14.75$_{\pm0.13}$ & 62.30$_{\pm3.76}$ & 31.65$_{\pm0.16}$ & 14.83$_{\pm0.03}$ & \underline{14.47$_{\pm0.02}$} & \cellcolor{gray!18}\textbf{14.45$_{\pm0.05}$} & 14.63$_{\pm0.08}$ & 14.73$_{\pm0.03}$ & 14.92$_{\pm0.07}$ & 16.93$_{\pm0.49}$ & 14.70$_{\pm0.07}$ & 14.81$_{\pm0.05}$ \\
 &  & 12 & 17.34$_{\pm0.66}$ & 16.44$_{\pm0.23}$ & 62.43$_{\pm3.69}$ & 33.09$_{\pm0.16}$ & 16.90$_{\pm0.05}$ & \cellcolor{gray!18}\textbf{16.21$_{\pm0.06}$} & 16.27$_{\pm0.11}$ & 16.51$_{\pm0.15}$ & \underline{16.25$_{\pm0.02}$} & 16.90$_{\pm0.04}$ & 18.56$_{\pm0.70}$ & 16.54$_{\pm0.09}$ & 16.86$_{\pm0.08}$ \\
 &  & Avg & 17.34$_{\pm0.66}$ & 16.44$_{\pm0.23}$ & 62.43$_{\pm3.69}$ & 33.09$_{\pm0.16}$ & 16.90$_{\pm0.05}$ & \cellcolor{gray!18}\textbf{16.21$_{\pm0.06}$} & 16.27$_{\pm0.11}$ & 16.51$_{\pm0.15}$ & \underline{16.25$_{\pm0.02}$} & 16.90$_{\pm0.04}$ & 18.56$_{\pm0.70}$ & 16.54$_{\pm0.09}$ & 16.86$_{\pm0.08}$ \\
\cmidrule(lr){2-16}
 & \multirow{4}{*}{MAPE (\%)} & 3 & 31.90$_{\pm3.28}$ & 28.72$_{\pm1.62}$ & 394.65$_{\pm50.28}$ & 173.05$_{\pm3.23}$ & 25.46$_{\pm0.86}$ & 25.02$_{\pm0.28}$ & \cellcolor{gray!18}\textbf{24.18$_{\pm0.51}$} & 25.92$_{\pm1.33}$ & 26.39$_{\pm1.24}$ & 25.80$_{\pm0.71}$ & 53.38$_{\pm6.05}$ & 26.09$_{\pm0.34}$ & \underline{24.84$_{\pm0.08}$} \\
 &  & 6 & 34.34$_{\pm3.42}$ & 30.33$_{\pm1.75}$ & 395.25$_{\pm50.66}$ & 173.67$_{\pm2.41}$ & 26.61$_{\pm0.79}$ & 26.21$_{\pm0.40}$ & \cellcolor{gray!18}\textbf{25.19$_{\pm0.48}$} & 26.91$_{\pm1.15}$ & 27.07$_{\pm1.10}$ & 27.17$_{\pm0.52}$ & 54.64$_{\pm6.06}$ & 26.97$_{\pm0.26}$ & \underline{26.07$_{\pm0.01}$} \\
 &  & 12 & 38.23$_{\pm3.32}$ & 34.57$_{\pm2.34}$ & 396.12$_{\pm50.73}$ & 178.47$_{\pm4.31}$ & 29.34$_{\pm0.78}$ & 29.24$_{\pm0.78}$ & \cellcolor{gray!18}\textbf{27.65$_{\pm0.36}$} & 29.83$_{\pm0.69}$ & \underline{28.75$_{\pm1.08}$} & 30.37$_{\pm0.21}$ & 56.99$_{\pm6.78}$ & 29.48$_{\pm0.45}$ & 28.89$_{\pm0.12}$ \\
 &  & Avg & 38.23$_{\pm3.32}$ & 34.57$_{\pm2.34}$ & 396.12$_{\pm50.73}$ & 178.47$_{\pm4.31}$ & 29.34$_{\pm0.78}$ & 29.24$_{\pm0.78}$ & \cellcolor{gray!18}\textbf{27.65$_{\pm0.36}$} & 29.83$_{\pm0.69}$ & \underline{28.75$_{\pm1.08}$} & 30.37$_{\pm0.21}$ & 56.99$_{\pm6.78}$ & 29.48$_{\pm0.45}$ & 28.89$_{\pm0.12}$ \\
\bottomrule
\end{tabular}%
}
\end{table}

\begin{table*}[!b]
\centering
\color{blue}
\caption{\revisiontext{Full-12-lead MAE of the seven methods evaluated in all nine districts. Entries are mean$\pm$sample standard deviation after equal-year aggregation over seeds 47--49, except on PEMS04, which uses seeds 47--51 as in Table~\ref{tab:tsas_full_horizon_summary}. All scores every accepted target, and New scores sensors new to the preceding year. Parentheses give the number of eligible years, and empty new-node cohorts are omitted. Every method except Pretrain has current-year training. Each district is trained and evaluated separately, so these comparisons are not cross-district transfer or zero-shot cold start. Bold and underline indicate the two lowest unrounded means in each row, without a significance claim. PEMS04 and PEMS05 also have the full 13-method comparison in Tables~\ref{tab:tsas_evo_part1} and~\ref{tab:tsas_new_sensors_part1}.}}
\label{tab:tsas_three_seed_diagnostics}
\resizebox{\textwidth}{!}{%
\begin{tabular}{cc|cccc|cc|c}
\toprule
\multirow{2}{*}{\textbf{District}} & \multirow{2}{*}{\textbf{Cohort (years)}} & \multicolumn{4}{c|}{\textbf{Training schemes}} & \multicolumn{2}{c|}{\textbf{Continual adaptations}} & \multicolumn{1}{c}{\textbf{Retrieval/calibration}} \\
\cmidrule(lr){3-6} \cmidrule(lr){7-8} \cmidrule(lr){9-9}
 & & \textbf{Pretrain} & \textbf{Retrain} & \textbf{Online-NN} & \textbf{Online-AN} & \textbf{TrafficStream} & \textbf{EAC} & \textbf{ST-TTC} \\
\midrule
\textbf{PEMS03} & All (25) & 26.86$_{\pm0.80}$ & 14.30$_{\pm0.08}$ & 16.83$_{\pm0.37}$ & \cellcolor{gray!18}\textbf{13.76$_{\pm0.10}$} & \underline{14.14$_{\pm0.01}$} & 14.58$_{\pm0.18}$ & 14.26$_{\pm0.04}$ \\
2001--2025 & New (21) & 20.56$_{\pm0.80}$ & 11.28$_{\pm0.05}$ & \underline{10.86$_{\pm0.05}$} & \cellcolor{gray!18}\textbf{10.83$_{\pm0.07}$} & 10.99$_{\pm0.01}$ & 11.37$_{\pm0.18}$ & 11.23$_{\pm0.04}$ \\
\midrule
\textbf{PEMS04} & All (24) & 42.10$_{\pm1.61}$ & 22.08$_{\pm0.09}$ & 23.74$_{\pm0.83}$ & \cellcolor{gray!18}\textbf{21.39$_{\pm0.09}$} & 22.01$_{\pm0.37}$ & 24.51$_{\pm2.60}$ & \underline{21.96$_{\pm0.09}$} \\
2002--2025 & New (20) & 41.51$_{\pm1.72}$ & 18.69$_{\pm0.07}$ & \cellcolor{gray!18}\textbf{17.80$_{\pm0.16}$} & \underline{18.02$_{\pm0.07}$} & 18.29$_{\pm0.35}$ & 22.31$_{\pm3.31}$ & 18.58$_{\pm0.07}$ \\
\midrule
\textbf{PEMS05} & All (21) & 27.44$_{\pm0.17}$ & 11.92$_{\pm0.05}$ & 13.96$_{\pm0.32}$ & \cellcolor{gray!18}\textbf{11.52$_{\pm0.04}$} & 13.71$_{\pm0.05}$ & 14.06$_{\pm0.59}$ & \underline{11.90$_{\pm0.04}$} \\
2005--2025 & New (12) & 24.95$_{\pm0.26}$ & 10.37$_{\pm0.05}$ & \underline{10.04$_{\pm0.06}$} & \cellcolor{gray!18}\textbf{9.98$_{\pm0.03}$} & 10.23$_{\pm0.05}$ & 12.44$_{\pm0.40}$ & 10.34$_{\pm0.05}$ \\
\midrule
\textbf{PEMS06} & All (21) & 36.51$_{\pm1.24}$ & 14.85$_{\pm0.10}$ & 15.18$_{\pm0.10}$ & \cellcolor{gray!18}\textbf{14.18$_{\pm0.04}$} & 15.40$_{\pm0.19}$ & 17.78$_{\pm0.69}$ & \underline{14.80$_{\pm0.12}$} \\
2005--2025 & New (17) & 36.44$_{\pm1.40}$ & 13.14$_{\pm0.01}$ & \underline{12.72$_{\pm0.09}$} & \cellcolor{gray!18}\textbf{12.63$_{\pm0.00}$} & 13.26$_{\pm0.11}$ & 15.75$_{\pm0.62}$ & 13.09$_{\pm0.04}$ \\
\midrule
\textbf{PEMS07} & All (25) & 39.96$_{\pm0.62}$ & 18.31$_{\pm0.05}$ & 20.58$_{\pm0.17}$ & \cellcolor{gray!18}\textbf{17.62$_{\pm0.09}$} & \underline{17.84$_{\pm0.11}$} & 17.93$_{\pm0.52}$ & 18.20$_{\pm0.04}$ \\
2001--2025 & New (19) & 41.81$_{\pm1.51}$ & 17.69$_{\pm0.06}$ & \cellcolor{gray!18}\textbf{16.37$_{\pm0.06}$} & 16.96$_{\pm0.04}$ & \underline{16.72$_{\pm0.09}$} & 17.09$_{\pm0.41}$ & 17.52$_{\pm0.03}$ \\
\midrule
\textbf{PEMS08} & All (25) & 27.93$_{\pm0.36}$ & 15.55$_{\pm0.05}$ & 16.00$_{\pm0.01}$ & \cellcolor{gray!18}\textbf{14.97$_{\pm0.07}$} & \underline{15.35$_{\pm0.03}$} & 15.99$_{\pm0.46}$ & 15.47$_{\pm0.05}$ \\
2001--2025 & New (21) & 28.53$_{\pm0.58}$ & 15.65$_{\pm0.04}$ & \cellcolor{gray!18}\textbf{14.93$_{\pm0.08}$} & \underline{15.05$_{\pm0.06}$} & 15.17$_{\pm0.02}$ & 15.71$_{\pm0.12}$ & 15.56$_{\pm0.04}$ \\
\midrule
\textbf{PEMS10} & All (20) & 61.01$_{\pm1.18}$ & 12.08$_{\pm0.03}$ & 12.22$_{\pm0.03}$ & \cellcolor{gray!18}\textbf{11.66$_{\pm0.06}$} & 12.06$_{\pm0.05}$ & 12.56$_{\pm0.14}$ & \underline{12.02$_{\pm0.06}$} \\
2006--2025 & New (19) & 64.27$_{\pm1.18}$ & 9.92$_{\pm0.02}$ & \cellcolor{gray!18}\textbf{9.53$_{\pm0.03}$} & \underline{9.60$_{\pm0.04}$} & 9.78$_{\pm0.05}$ & 10.73$_{\pm0.10}$ & 9.85$_{\pm0.06}$ \\
\midrule
\textbf{PEMS11} & All (27) & 42.44$_{\pm2.08}$ & 18.37$_{\pm0.02}$ & 19.31$_{\pm0.18}$ & \cellcolor{gray!18}\textbf{17.64$_{\pm0.03}$} & 18.83$_{\pm0.67}$ & 20.63$_{\pm0.42}$ & \underline{18.22$_{\pm0.02}$} \\
1999--2025 & New (22) & 42.51$_{\pm1.88}$ & 18.42$_{\pm0.04}$ & \cellcolor{gray!18}\textbf{17.67$_{\pm0.22}$} & \underline{17.75$_{\pm0.04}$} & 18.97$_{\pm0.79}$ & 21.04$_{\pm0.60}$ & 18.27$_{\pm0.04}$ \\
\midrule
\textbf{PEMS12} & All (24) & 19.84$_{\pm0.38}$ & 15.88$_{\pm0.01}$ & 15.96$_{\pm0.15}$ & \underline{15.25$_{\pm0.10}$} & 15.50$_{\pm0.10}$ & \cellcolor{gray!18}\textbf{15.12$_{\pm0.22}$} & 15.82$_{\pm0.02}$ \\
2002--2025 & New (18) & 19.57$_{\pm0.29}$ & 15.39$_{\pm0.04}$ & \cellcolor{gray!18}\textbf{14.40$_{\pm0.10}$} & 14.78$_{\pm0.12}$ & \underline{14.69$_{\pm0.07}$} & 14.74$_{\pm0.24}$ & 15.32$_{\pm0.04}$ \\
\bottomrule
\end{tabular}%
}
\end{table*}

\begin{table}[t]
\centering
\color{blue}
\small
\caption{\revisiontext{Catalog coverage of a retrospective fixed intersection. Full-span counts use each district's evaluated period, and the recent intersection uses 2018--2025. Counts describe catalog membership, not continuous valid measurements or forecasting accuracy.}}
\label{tab:fixed_catalog_coverage}
\begin{tabular}{lrrr}
\toprule
District & Full-span common & Recent common & 2025 catalog \\
\midrule
PEMS03 & 130 & 909 & 1,859 \\
PEMS04 & 703 & 3,752 & 4,110 \\
PEMS05 & 0 & 240 & 572 \\
PEMS06 & 0 & 476 & 746 \\
PEMS10 & 44 & 1,003 & 1,396 \\
\bottomrule
\end{tabular}
\end{table}

\begin{table}[t]
\centering
\color{blue}
\small
\caption{\revisiontext{PEMS04 Online-AN coordinate-policy sensitivity: mean $\pm$ sample standard deviation over matched seeds 47--49. Twelve-lead metrics first average equally over eligible years (24 for all sensors, 20 for new-to-previous-year sensors). The retrospective column is the matching three-seed subset, not the main five-seed mean. Relative changes compare the means. Lower errors are better.}}
\label{tab:tsas_coordinate_sensitivity}
\begin{tabular}{llrrr}
\toprule
Cohort & Metric & Retrospective & Dated coordinates/isolation & Change \\
\midrule
All & MAE & $21.346\pm0.083$ & $21.926\pm0.139$ & $+2.72\%$ \\
All & RMSE & $34.986\pm0.140$ & $35.770\pm0.220$ & $+2.24\%$ \\
All & MAPE (\%) & $19.595\pm0.342$ & $20.152\pm0.463$ & $+2.84\%$ \\
New-to-previous & MAE & $17.988\pm0.078$ & $20.038\pm0.225$ & $+11.40\%$ \\
New-to-previous & RMSE & $30.233\pm0.111$ & $33.313\pm0.286$ & $+10.19\%$ \\
New-to-previous & MAPE (\%) & $23.496\pm0.997$ & $22.918\pm0.484$ & $-2.46\%$ \\
\bottomrule
\end{tabular}
\end{table}

\begin{table}[t]
\centering
\color{blue}
\small
\caption{\revisiontext{Training-free references on the corrected yearly tasks: the training mean of each year predicted for every target, and the last input value repeated for all 12 leads. Same accepted targets, cumulative horizons, and equal-year aggregation as the method results. MAPE is in percent. These deterministic references have no seed variation.}}
\label{tab:tsas_naive_references}
\setlength{\tabcolsep}{3.5pt}
\begin{tabular}{lll|rrr|rrr|rrr}
\toprule
 & & & \multicolumn{3}{c|}{MAE} & \multicolumn{3}{c|}{RMSE} & \multicolumn{3}{c}{MAPE (\%)} \\
District & Cohort & Reference & 3 & 6 & 12 & 3 & 6 & 12 & 3 & 6 & 12 \\
\midrule
PEMS04 & All & Training mean & 134.77 & 134.78 & 134.81 & 162.76 & 162.76 & 162.77 & 449.48 & 449.78 & 450.46 \\
PEMS04 & All & Last value & 19.35 & 22.06 & 27.75 & 32.05 & 36.28 & 45.18 & 16.80 & 18.71 & 22.96 \\
PEMS04 & New & Training mean & 138.08 & 138.08 & 138.09 & 162.25 & 162.25 & 162.26 & 674.41 & 674.84 & 675.85 \\
PEMS04 & New & Last value & 16.01 & 18.46 & 23.54 & 27.37 & 31.23 & 39.42 & 19.13 & 21.02 & 25.30 \\
\midrule
PEMS05 & All & Training mean & 61.27 & 61.28 & 61.32 & 77.35 & 77.35 & 77.37 & 396.44 & 396.73 & 397.41 \\
PEMS05 & All & Last value & 11.65 & 12.79 & 15.27 & 18.75 & 20.71 & 24.89 & 24.50 & 26.49 & 31.18 \\
PEMS05 & New & Training mean & 60.05 & 60.07 & 60.11 & 70.50 & 70.51 & 70.54 & 541.58 & 542.08 & 543.13 \\
PEMS05 & New & Last value & 10.18 & 11.14 & 13.24 & 16.21 & 17.85 & 21.37 & 28.90 & 30.84 & 35.51 \\
\bottomrule
\end{tabular}
\end{table}

\begin{table}[t]
\centering
\color{blue}
\small
\caption{\revisiontext{Full-12-lead MAE over the 2018--2025 tasks only, taken from the same sequential runs as Table~\ref{tab:tsas_full_horizon_summary}. Each seed weights the valid years of this period equally (8 all-sensor and 7 new-to-previous-year years for PEMS04, 8 and 8 for PEMS05). Entries are mean$\pm$sample standard deviation over seeds 47--51 (PEMS04) or 47--49 (PEMS05). Each method retains its original reset or state-carryover policy. Bold marks the lowest mean in each column, and differences are descriptive.}}
\label{tab:tsas_recent_slice}
\setlength{\tabcolsep}{5pt}
\begin{tabular}{lrrrr}
\toprule
 & \multicolumn{2}{c}{PEMS04, 2018--2025} & \multicolumn{2}{c}{PEMS05, 2018--2025} \\
\cmidrule(lr){2-3}\cmidrule(lr){4-5}
Method & All & New & All & New \\
\midrule
DCRNN & $20.355\pm0.614$ & $16.734\pm0.798$ & $11.456\pm0.163$ & $10.316\pm0.299$ \\
ASTGCN & $20.436\pm0.228$ & $16.899\pm0.235$ & $10.835\pm0.301$ & $9.707\pm0.254$ \\
TGCN & $131.992\pm0.010$ & $136.612\pm0.064$ & $58.069\pm1.856$ & $55.046\pm0.795$ \\
Pretrain & $44.844\pm3.718$ & $46.335\pm4.357$ & $28.177\pm0.173$ & $26.070\pm0.291$ \\
Retrain & $19.081\pm0.055$ & $15.190\pm0.112$ & $10.730\pm0.037$ & $9.601\pm0.041$ \\
Online-NN & $19.442\pm0.126$ & $14.051\pm0.086$ & $10.980\pm0.037$ & $9.239\pm0.062$ \\
Online-AN & $18.463\pm0.048$ & $14.539\pm0.047$ & $\mathbf{10.275\pm0.017}$ & $9.207\pm0.039$ \\
TrafficStream & $18.778\pm0.218$ & $14.516\pm0.260$ & $10.642\pm0.051$ & $9.421\pm0.046$ \\
PECPM & $\mathbf{17.815\pm0.057}$ & $\mathbf{13.978\pm0.116}$ & $10.278\pm0.076$ & $\mathbf{9.069\pm0.139}$ \\
STKEC & $19.035\pm0.118$ & $15.060\pm0.161$ & $10.759\pm0.031$ & $9.526\pm0.032$ \\
EAC & $21.237\pm3.727$ & $18.003\pm4.539$ & $12.787\pm0.576$ & $11.599\pm0.448$ \\
STRAP & $18.598\pm0.038$ & $14.654\pm0.067$ & $10.530\pm0.020$ & $9.421\pm0.032$ \\
ST-TTC & $18.985\pm0.055$ & $15.089\pm0.125$ & $10.698\pm0.093$ & $9.593\pm0.086$ \\
\bottomrule
\end{tabular}
\end{table}

\begin{table}[t]
\centering
\color{blue}
\small
\caption{\revisiontext{Fixed- versus evolving-catalog training on the sensors present in every 2018--2025 catalog ($S$: 240 of the 572 PEMS05 and 476 of the 746 PEMS06 sensors of 2025). E trains on each year's full catalog and F on $S$ only. Both start new models in 2018 and are scored on the same accepted test targets of $S$ over the full 12-lead horizon. Entries are mean$\pm$sample standard deviation over seeds 47--49 of equal-year 2018--2025 means. E$-$F is paired within each seed, so a negative value means a lower error with evolving-catalog training. MAPE is in percent and its difference in percentage points.}}
\label{tab:tsas_fixed_vs_evolving}
\setlength{\tabcolsep}{5pt}
\begin{tabular}{lllrrr}
\toprule
District & Method & Metric & E & F & E$-$F \\
\midrule
PEMS05 & Retrain & MAE & $12.705\pm0.043$ & $12.800\pm0.045$ & $-0.096\pm0.041$ \\
 &  & RMSE & $20.396\pm0.063$ & $20.435\pm0.079$ & $-0.039\pm0.083$ \\
 &  & MAPE & $23.310\pm0.530$ & $24.268\pm0.535$ & $-0.958\pm0.963$ \\
 & Online-AN & MAE & $12.452\pm0.139$ & $12.539\pm0.069$ & $-0.087\pm0.180$ \\
 &  & RMSE & $19.882\pm0.303$ & $19.949\pm0.180$ & $-0.067\pm0.378$ \\
 &  & MAPE & $22.836\pm0.233$ & $23.793\pm0.623$ & $-0.957\pm0.482$ \\
\midrule
PEMS06 & Retrain & MAE & $13.253\pm0.124$ & $13.228\pm0.178$ & $+0.025\pm0.063$ \\
 &  & RMSE & $22.545\pm0.195$ & $22.448\pm0.290$ & $+0.097\pm0.115$ \\
 &  & MAPE & $24.726\pm0.785$ & $24.623\pm1.292$ & $+0.103\pm0.534$ \\
 & Online-AN & MAE & $12.992\pm0.029$ & $12.882\pm0.144$ & $+0.110\pm0.115$ \\
 &  & RMSE & $22.027\pm0.078$ & $21.744\pm0.281$ & $+0.283\pm0.310$ \\
 &  & MAPE & $23.778\pm0.550$ & $23.662\pm0.424$ & $+0.115\pm0.310$ \\
\bottomrule
\end{tabular}
\end{table}

\end{document}